%% file: AAAI23_camera.tex
\documentclass[letterpaper]{article} 
\usepackage{aaai23}  
\usepackage{times}  
\usepackage{helvet}  
\usepackage{courier}  
\usepackage[hyphens]{url}  
\usepackage{graphicx} 
\usepackage{natbib}  
\usepackage{caption} 
\input{preamble}

\usepackage{algorithm}
\usepackage{algorithmic}

\usepackage{newfloat}
\usepackage{listings}
\DeclareCaptionStyle{ruled}{labelfont=normalfont,labelsep=colon,strut=off} 
\floatstyle{ruled}
\newfloat{listing}{tb}{lst}{}
\floatname{listing}{Listing}
\title{Task and Model Agnostic Adversarial Attack on Graph Neural Networks}
\author {
    Kartik Sharma \textsuperscript{\rm 1}\thanks{Work done as undergraduate thesis at IIT-Delhi},
    Samidha Verma \textsuperscript{\rm 2},
    Sourav Medya \textsuperscript{\rm 3}, \\
    Arnab Bhattacharya \textsuperscript{\rm 4},
    Sayan Ranu \textsuperscript{\rm 2}
}
\affiliations {
    \textsuperscript{\rm 1} Georgia Institute of Technology, Atlanta, USA\\
    \textsuperscript{\rm 2} Indian Institute of Technology, Delhi, India\\
    \textsuperscript{\rm 3} University of Illinois, Chicago, USA\\
    \textsuperscript{\rm 4} Indian Institute of Technology, Kanpur, India\\
    ksartik@gatech.edu, csy207575@iitd.ac.in, medya@uic.edu, sayanranu@iitd.ac.in, arnabb@iitk.ac.in
}

\begin{document}

\maketitle
\input{0_abstract}

\input{1_introduction}
\input{3_prob}
\input{4_approach}

\input{5_experiment}

\input{7_conclusion}
\newpage

\section*{Acknowledgements}
We thank the HPC facility of IIT Delhi for computational resources and the reviewers for their constructive feedback. 

\bibliography{citations}
\clearpage
\appendix
\input{8_appendix}

\end{document}

%% file: preamble.tex
\usepackage{soul}
\usepackage{url}
\usepackage{epstopdf}
\usepackage[utf8]{inputenc}
\usepackage{amsmath}
\usepackage{amssymb}
\usepackage{booktabs}
\usepackage{algorithm}
\usepackage[noend]{algorithmic}
\usepackage{comment}
\usepackage{pifont}
\usepackage{smartdiagram}
\usepackage{multirow}
\usepackage{diagbox}

\usepackage{booktabs} 
\usepackage{array}
\usepackage{color,soul,xspace}

\usepackage{comment}
\usepackage{epsfig}
\usepackage{subfig}
\usepackage{mathrsfs,mdwlist,enumitem}
\usepackage{caption}
\newcommand{\ks}[1]{\textcolor{red}{[KS: #1]}}

\newcommand{\seer}{\texttt{CiteSeer}\xspace}
\newcommand{\cora}{\texttt{CoRA}\xspace}

\newcommand{\pubmed}{\texttt{PubMed}\xspace}

\newcommand{\dqn}{\textsc{DQN}\xspace}
\newcommand{\dai}{\textsc{RL-S2V}\xspace}
\newcommand{\gfattack}{\textsc{GF-Attack}\xspace}
\newcommand{\name}{\textsc{Tandis}\xspace}
\newcommand{\random}{\textsc{Random}\xspace}
\newcommand{\degree}{\textsc{Degree}\xspace}

\newcommand{\gnns}{\textsc{Gnn}s\xspace}
\newcommand{\deepwalk}{\textsc{DeepWalk}\xspace}

\newcommand{\gnn}{\textsc{Gnn}\xspace}
\newcommand{\gcn}{\textsc{Gcn}\xspace}
\newcommand{\gat}{\textsc{Gat}\xspace}
\newcommand{\gin}{\textsc{Gin}\xspace}
\newcommand{\guard}{\textsc{Gnn-Guard}\xspace}
\newcommand{\gsage}{\textsc{GraphSage}\xspace}
\newcommand{\pgnn}{\textsc{P-Gnn}\xspace}
\newcommand{\CG}{\mathcal{G}\xspace}
\newcommand{\CM}{\mathcal{M}\xspace}
\newcommand{\CN}{\mathcal{N}\xspace}
\newcommand{\CB}{\mathcal{B}\xspace}
\newcommand{\CP}{\mathcal{P}\xspace}
\newcommand{\CV}{\mathcal{V}\xspace}
\newcommand{\CE}{\mathcal{E}\xspace}
\newcommand{\CC}{\mathcal{C}\xspace}
\newcommand{\CX}{\boldsymbol{\mathcal{X}\xspace}}

\newcommand{\CZ}{\mathbf{Z}\xspace}

\newcommand{\CA}{\boldsymbol{\mathcal{A}}\xspace}

\newcommand{\SM}{\mathscr{M}\xspace}
\newcommand{\cW}{\mathbf{W}\xspace}
\newcommand{\cw}{\mathbf{w}\xspace}
\newcommand{\CL}{\mathcal{L}\xspace}

\newcommand{\ch}{\mathbf{h}\xspace}

\newcommand{\cz}{\mathbf{z}\xspace}

\setlist{nolistsep,leftmargin=*}

\newcommand{\SC}{{\sc Set Cover}\xspace}

\newtheorem{thm}{\textbf{Theorem}}

\newtheorem{definition}{\textbf{Definition}}

\newtheorem{prob}{\textbf{Problem}}


%% file: 0_abstract.tex
\begin{abstract}
Adversarial attacks on Graph Neural Networks (\gnns) reveal their security vulnerabilities, limiting their adoption in safety-critical applications. However, existing attack strategies rely on the knowledge of either the \gnn model being used or the predictive task being attacked. \textit{Is this knowledge necessary?} For example, a graph may be used for multiple downstream tasks unknown to a practical attacker. It is thus important to test the vulnerability of \gnns to adversarial perturbations in a model and task agnostic setting. In this work, we study this problem and show that \gnns remain vulnerable even when the downstream task and model are unknown. The proposed algorithm, \name (\underline{\textbf{T}}argeted \underline{\textbf{A}}ttack via \underline{\textbf{N}}eighborhood \underline{\textbf{DIS}}tortion) shows that distortion of node neighborhoods is effective in drastically compromising prediction performance. Although neighborhood distortion is an NP-hard problem, \name designs an effective heuristic through a novel combination of \textit{Graph Isomorphism Network} with \textit{deep $Q$-learning}. Extensive experiments on real datasets and state-of-the-art models show that, on average, \name is up to $50\%$ more effective than state-of-the-art techniques, while being more than $1000$ times faster.
\end{abstract}

%% file: 1_introduction.tex
\section{Introduction and Related work}
\label{sec:intro}
Graph neural networks (\gnns) ~\cite{hamilton2017graphsage,kipf2017iclr,iclr2018gat,graphreach,lgnn}, have received much success in structural prediction tasks \cite{graphgen,greed,physics,neuromlr,gcomb}. 
Consequently, \gnns have witnessed significant adoption in the industry~\cite{pinsage,uber}. It is therefore imperative to ensure that \gnns are secure and robust to adversarial attacks. In this work, we investigate this aspect, identify vulnerabilities, and link them to graph properties that potentially hold the solution towards making \gnns more secure.

An attack may occur either during training (\textit{poisoning}) or testing (\textit{evasion}). 
The attack strategy depends on the extent of information access. In the literature, primarily, three modes of information access has been studied. 

\begin{enumerate}
\item \textbf{White-box attack (WBA):} In a white-box attack \cite{liu2019unified,wang2019acm,wu2019ijcai,xu2019ijcai}, the attacker has access to all information such as model architecture, model parameters and training hyperparameters. 


\item \textbf{Black-box attack (BBA):} In a black-box attack \cite{bojchevski2019adversarial,chang2020restricted,chen2021graphattacker,dai2018adversarial,gupta2021adversarial,li2020adversarial,ma2020towards,ma2019attacking,wang2020efficient}, the attacker does not have access to the training model parameters. The attacker can only pose black-box queries on a test sample and get the model output in return (such as a label or score). 


\item \textbf{Grey-box attack (GBA):} A grey-box attack is a mix of white-box and black-box attacks \cite{liu2019unified,sun2020www,wang2019acm,zugner2018adversarial,zugner_adversarial_2019}, where the attacker has partial information of the training model. The extent of partial information depends on the particular attack.
\end{enumerate}
In this work, we focus on \textit{targeted black-box, evasion attacks}~\cite{dai2018adversarial}. Specifically, we consider an input test graph $\CG=(\CV,\CE, \CX)$, a target node $t\in\CV$ and a budget $\CB$. Our goal is to introduce at most $\CB$ edge alterations (additions or deletions) in $\CG$ such that the performance of an \textit{unknown} \gnn model $\CM$ is maximally reduced on node $t$. 

\noindent
\textbf{Limitations of Existing Works:}
Although a black-box attack does not require any information on the model parameters, they operate under three key assumptions:
\begin{enumerate}
\item \textbf{Task-specific strategy:} Most \gnn models are trained for a specific task (e.g., node classification) using an appropriately chosen loss function. Existing techniques for adversarial attacks tailor their strategy towards a specific prediction task under the assumption that this task is known. Hence, they do not generalize to unseen tasks. While this assumption acts as an added layer of security for \gnn models, we ask the question: \textit{Is it possible to design task-agnostic adversarial attacks?}


\item \textbf{Knowledge of the \gnn model:} Although black-box attacks do not need knowledge of the model parameters, they often assume the model type. For example, an attack may be customized for \gcn and, therefore, rendered ineffective if the training model switches to Locality-aware \gnns~\cite{you2019pgnn,graphreach}. 

\item \textbf{Label-dependent:} Several BBA algorithms require knowledge of the ground-truth data. As an example, an algorithm may require knowledge of node labels to adversarially attack node classification. These ground truth data is often not available in public domain. For example, Facebook may tag users based on their primary interest area. But this information is proprietary.  

\end{enumerate}


\noindent
\textbf{Contributions:}
Table~\ref{tab:summary} summarizes the limitations in existing algorithms for targeted BBA. In this work, we bridge this gap. Our key contributions are as follows:
\begin{itemize}
    \item We formulate the problem of \textit{task, model, and label agnostic black-box evasion attacks} on \gnns. The proposed algorithm, \name (\underline{\textbf{T}}argeted \underline{\textbf{A}}ttack via \underline{\textbf{N}}eighborhood \underline{\textbf{DIS}}tortion), shows that such attacks are indeed possible.
    
    
    \item \name exploits the observation that, regardless of the task or the model-type, if the neighborhood of a node can be \textit{distorted}, the downstream prediction task would be affected. Our analysis shows that budget-constrained neighborhood distortion is NP-hard. This computational bottleneck is overcome, by using a \textit{Graph Isomorphism Network} (\gin) to embed neighborhoods and then using \textit{deep $Q$-learning} (\dqn) to explore this combinatorial space in an effective and efficient manner. 
    
    \item Extensive experiments on real datasets show that, on average, \name is up to $50\%$ more effective than state-of-the-art techniques for BBA evasion attacks and $1000$ times faster. More importantly, \name establishes that task and model agnosticism do not impart any additional security to \gnns. 
\end{itemize}

\begin{table}[t]
    \centering
    \scalebox{0.9}{
    \begin{tabular}{cccccc}
    \toprule
    \textbf{Algorithm}   & \textbf{Task} & \textbf{Model} &\textbf{Label} \\
      \midrule
      \dai\cite{dai2018adversarial} &  &\checkmark & \\
      \gfattack\cite{chang2020restricted} & \checkmark & &\checkmark \\
      Wang et al.~\cite{wang2020efficient} &  & \checkmark& \\
       \textbf{ \name} & \checkmark&\checkmark&\checkmark\\
     \bottomrule
    \end{tabular}}
    \caption{\textbf{Characterization of existing targeted, black-box evasion attacks based on agnosticism of the features.}}
    \label{tab:summary}
\end{table}

%% file: 3_prob.tex
\section{Problem Formulation}
\label{sec:formulation}
We denote a graph as $\CG=(\CV,\CE,\CX)$ where $\CV$ is the set of nodes, $\CE$ the set of edges, and $\CX$ is the set of attribute vectors corresponding to each node $v\in\CV$. In this paper, we assume $\CG$ to be an undirected graph. Nonetheless, all of the proposed methodologies easily extend to directed graphs. The \textit{distance} $sp(v,u)$ between nodes $v$ and $u$ is measured in the terms of the length of the shortest path from $v$ to $u$. 
The $k$-hop neighborhood of a node $v$ is therefore defined as $\CN^k_{\CG}(v)=\{u\in\CV \mid sp(v,u)\leq k\}$. 

The adjacency matrix of graph $\CG$ is denoted as $\CA_{\CG}$. The \textit{distance} between two undirected graphs, $\CG=(\CG,\CE,\CX)$ and $\CG'=(\CV,\CE',\CX)$, with the same node set but different edge set, is defined to be half of the $L_1$ distance between their adjacency matrices:
$d(\CG,\CG')=\|\CA_{\CG}-\CA_{\CG'}\|/2$.

Given a \gnn model $\CM$, its performance on graph $\CG$ towards a particular predictive task is quantified using a \textit{performance metric} $\CP_{\CM}(\CG)$:  
\begin{equation}
\label{eq:loss1}
    \CP^*_{\CM}(\CG)=f_1\left(\{\CP_{\CM}(\CG,v)\mid v\in\CV\}\right) \text{ , where }
\end{equation}
\vspace{-0.20in}
\begin{equation}
\nonumber
   \CP_{\CM}(\CG,v)=f_2(\cz_{v,\CG},\ell_v)
\end{equation}
The performance metric $\CP_{\CM}(\CG,v)$ on node $v$ is a function $f_2$ of its embedding $\cz_{v,\CG}$ and its ground truth label or score $\ell_v$. The overall performance $\CP^*_{\CM}(\CG)$ is some aggregation $f_1$ over the performance across all nodes. We will discuss some specific instances in Section~\ref{sec:experiments}.

An adversarial attacker wishes to change $\CG$ by performing $\CB$ edge deletions and additions so that the performance on a \textit{target node} $t$ is minimized. We assume that the attacker has access to a subset of nodes $\CC\subseteq\CV$ and edges may be modified only among the set $\CC\cup \{t\}$. Semantically, $\CC$ may represent colluding bots, or vulnerable users who have public profiles instead of private profiles, etc. Note that we do not explicitly allow node deletions and additions, since these can be modeled through deletions and additions over the edges~\cite{dai2018adversarial}. 
\begin{prob}[Targeted Black-box Evasion Attack]
\label{prob:prob1}
 Given graph $\CG=(\CV,\CE,\CX)$, a target node $t$, budget $\CB$, and a subset of accessible nodes $\CC\subseteq\CV$,  perform at most $\CB$ edge additions or deletions from edge space $\{(u,v)\mid u,v\in\CC\cup\{t\}\}$ to form graph $\CG^*$ such that 
\begin{equation}
\label{eq:problem}
\CG^*=\arg\min_{\CG': d(\CG,\CG') \leq \CB} \CP_{\CM}(\CG',t) 
\end{equation}
\end{prob}
\vspace{-0.05in}
Both the \gnn model $\CM$ and performance metric $\CP_{\CM}(\CG',t)$ are \textit{unknown} to the attacker. Note that our formulation is easily extensible to a set of target nodes, where Eq.~\ref{eq:problem} is aggregated over all targets.

\section{\name: Targeted Attack via Neighborhood DIStortion}
\label{sec:prob}
Since in the search process for $\CG^*$, the attacker does not have access to either the performance metric $\CP^*_{\CM}(\CG)$ or the model-type of $\CM$, we need a \textit{surrogate} function $\phi(\CG,t)$ in the \textit{graph space}, such that if $distance(\phi(\CG,t),\phi(\CG',t))\gg 0$ then $\CP_{\CM}(\CG,t)$ is significantly different from $\CP_{\CM}(\CG',t)$. 

\subsection{Neighbourhood Distortion in Graph Space}
\label{sec:graphDist}
To identify $\phi(\CG,t)$, we first note that there are primarily two types of \gnns: 

$\bullet$  \textbf{Neighborhood-convolution:} Examples of neighborhood convolution based architectures include \gcn~\cite{kipf2017iclr}, GraphSage~\cite{hamilton2017graphsage}, and \gat~\cite{iclr2018gat}. Here, the embedding of a node $v$ is computed through a \textit{convolution} operation $\psi$ over its $k$-hop neighborhood, i,e, $\cz_{v,\CG}=\psi\left(\CN^k_{\CG}(v)\right)$. Consequently, nodes with similar neighborhoods have similar embeddings~\cite{xu2019gin,you2019pgnn}. 

$\bullet$ \textbf{Locality-aware:} In locality-aware \gnns  such as \pgnn~\cite{you2019pgnn}, \textsc{GraphReach} \cite{graphreach}, \deepwalk \cite{perozzi2014deepwalk}, etc., two nodes have similar embeddings if they are located close to each other in the graph. 
Many real-world graphs display \textit{small-world} and \textit{scale-free} properties~\cite{albert2002statistical}. Both small-world and scale-free graphs usually have high clustering coefficients, which indicate that one-hop neighbors share many other neighbors as well. Thus, nearby nodes  have a high likelihood of having similar neighborhoods.

The above discussion reveals that node embeddings reflect node neighborhoods in both architectural paradigms. Consequently, the surrogate function $\phi(\CG,t)$ may be set to target node $t$'s $k$-hop neighborhood $\CN^k_{\CG}(t)$. If $\CN^k_{\CG}(t)$ can be distorted through adversarial edge edits, then it would also perturb the embedding of $t$, which in turn would affect the performance metric. Towards that end, we define the notion of \textit{neighborhood distortion} as follows.
\begin{definition}[Neighborhood Distortion]
Let $\mathcal{N}^k_{\CG}(v)$ and $\mathcal{N}^k_{\CG'}(v)$ be the $k$-hop neighborhoods of $v$ in original graph $\CG$ and adversarially perturbed graph $\CG'$ respectively. The distortion in $\CG'$ with respect to a node $v$ is simply the Jaccard distance of their neighborhoods.
\begin{equation}
    \delta(v,\CG,\CG')=1-\frac{|\mathcal{N}^k_{\CG}(v)\cap \mathcal{N}^k_{\CG'}(v)|}{|\mathcal{N}^k_{\CG}(v)\cup \mathcal{N}^k_{\CG'}(v)|}
\end{equation}
\end{definition}
Problem~\ref{prob:prob1} may now be reduced to the following objective:
\begin{equation}
\label{eq:nphard}
    \CG^*= \arg \max_{\CG': d(\CG,\CG')\leq \CB}{\delta(t,\CG,\CG')}  
\end{equation} 
where $t$ is the target node. Unfortunately, maximizing $\delta(t,\CG,\CG')$ is NP-hard even for $k=2$. 
\begin{thm}
\label{thm:nphard}
The problem $\max_{\CG': d(\CG,\CG')\leq \CB}{\delta(t,\CG,\CG')}$ is NP-hard and the objective is non-monotone and non-submodular. 
\end{thm}
\textsc{Proof.} We reduce the \SC problem to our maximization objective. 
See Appendix ~\ref{app:nphard} in the full version for more details~\cite{sharma2021task}. 
Owing to NP-hardness, computing the optimal solution for the problem in Eq.~\ref{eq:nphard} is not feasible. Furthermore, since the objective is neither submodular nor monotone, even the greedy algorithms that provide $1-1/e$ approximations are not applicable here \cite{nemhauser1978best}. We, therefore, investigate other optimization strategies. 

\begin{figure}[t]
    \centering
    \includegraphics[scale=0.5]{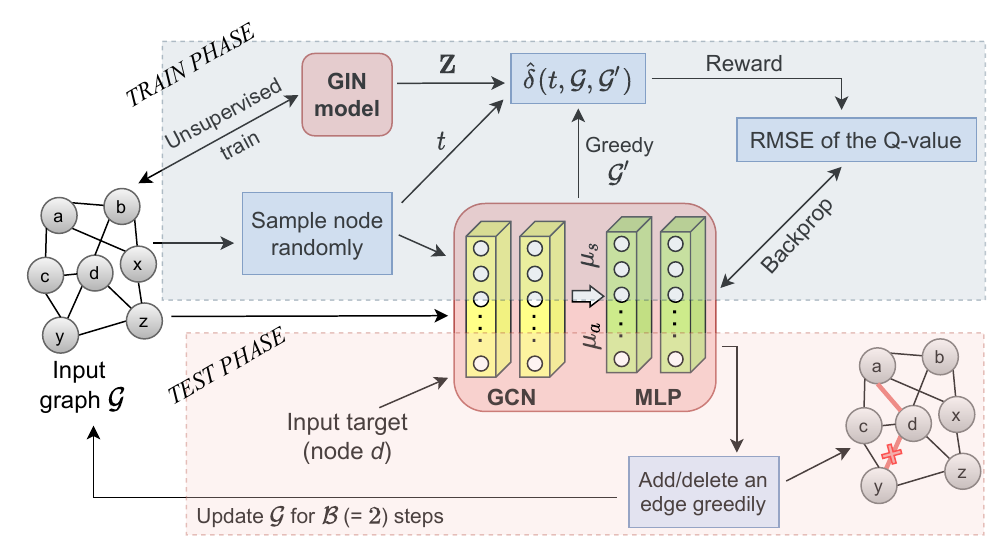}
  \caption{\textbf{Pipeline of \name in  train and test phase.}}
    \label{fig:flow_diag}
\end{figure}

%% file: 4_approach.tex

\subsection{A Reinforcement Learning Approach for Neighborhood Distortion}
\label{sec:rl}
Fig.~\ref{fig:flow_diag} presents the pipeline of our proposed algorithm \name. It proceeds through three phases.
\begin{enumerate}
    \item \textbf{Training data generation:} We capture a feature-space representation of node neighborhoods using \textit{Graph Isomorphism Network} (\gin)~\cite{xu2019gin}. This design choice is motivated by the property that \gin is as powerful as the \textit{Weisfeiler-Lehman graph isomorphism test}~\cite{xu2019gin}. On the feature space generated by \gin, we define a neighborhood distortion measure. 
    \item \textbf{Train phase:} We train a deep $Q$-learning network (\dqn), to predict the impact of an edge edit on distortion in the neighborhood embedding space. The proposed \dqn framework captures the \textit{combinatorial} nature of the problem, and shares \textit{parameters across budgets}.
    \item \textbf{Test phase:} Given an unseen graph, we perform  $\CB$ forward passes through the trained \dqn in an iterative manner to form the answer set of $\CB$ edge edits.
\end{enumerate}
\subsection{Training Data Generation}
\label{sec:train_gen}
The input to the training phase is a set of tuples of the form $\langle t, e, \SM(\CG),\SM(\CG_e) \rangle$. Here, $t$ is the target node, $e$ denotes an edge perturbation, and $\SM(\CG),\;\SM(\CG_e)$ are the \gin embeddings of the original graph $\CG$ and the graph formed following edge edit $e$, $\CG_e$, respectively. $\SM(\CG)=\{\CZ^{\CG}_v \mid v\in \CV\}$ contains the \gin embedding of each node in $\CG$. $\CZ^{\CG}_v$ characterizes the $k$-hop neighborhood of node $v$ in graph $\CG$. $t$ is randomly selected. The edge perturbations are selected using a greedy mechanism, which we detail in Appendix~\ref{app:greedy} of the full version~\cite{sharma2021task}.

\textbf{Embeddings via \gin:}
\gin draws its expressive power by deploying an \textit{injective} aggregation function. Specifically, in the initial layer, each node is characterized by an \textit{one-hot} vector of dimension $|\CV|$, which encodes the node ID. Subsequently, in each hidden layer $i$, we learn an embedding through the following transformation.
\begin{equation}
\label{eq:gin}
    \ch^{\CG}_{v,i}=MLP\left((1+\epsilon^i)\cdot\ch^{\CG}_{v,i-1}+\sum_{u\in \CN_{\CG}^1(v)}\ch^{\CG}_{u,i-1}\right)
\end{equation}

Here, $\epsilon^i$ is a layer-specific learnable parameter. The final embedding $\CZ_v^{\CG}=\ch^{\CG}_{v,k}$, where $k$ is final hidden layer. The embeddings are learned by minimizing the following \textit{unsupervised} loss function.
\begin{alignat}{1}
    \label{eq:loss}
    \CL(\SM(\CG))&=-\log \left(P\left(\CN_{\CG}^k(v)\mid \CZ_v^{\CG}\right)\right)\text{, where }\\ P\left(\CN_{\CG}^k(v)\mid \CZ_v^{\CG}\right)&=\prod_{u\in \CN_{\CG}^k(v)} P\left(u\mid \CZ_v^{\CG}\right) \text{ and }  
\end{alignat}
\begin{equation}
    P\left(u\mid \CZ_v^{\CG}\right) =\frac{exp\left(\CZ_u^{\CG}\cdot \CZ_v^{\CG}\right)}{\sum_{\forall u'\in\CV}exp\left(\CZ_{u'}^{\CG}\cdot\CZ_v^{\CG}\right)}
\end{equation}
For the pseudocode, please refer to Appendix~\ref{app:gin} of \citet{sharma2021task}.

\textbf{Distortion in embedding space:} 
We define distortion in the perturbed graph $\CG'$ with respect to the original graph $\CG$ as follows:
 \begin{equation} 
 \label{eq:silh}
 {\widehat{\delta}(t, \CG, \CG')} = {d_o(t) - d_p(t)} 
\end{equation}
Here, $d_o(t) = \frac{1}{\left\lvert\mathcal{N}^k_{\CG}(t)\right \rvert} \sum_{u \in \mathcal{N}^k_{\CG}(t)}{\left\lVert\CZ_t^{\CG'}- \CZ_u^{\CG'}\right\rVert_2}$ denotes the \textit{mean L2 distance} in the \textit{perturbed embedding} space between target node $t$ and the nodes in its original unperturbed neighborhood. On the other hand, $d_p(t) = \frac{1}{\left \lvert\mathcal{N}^k_{\CG'}(t)\right\rvert} \sum_{u \in \mathcal{N}^k_{\CG'}(t)}{\left\lVert\CZ_t^{\CG'}- \CZ_u^{\CG'}\right\rVert_2}$ denotes the same between target node $t$ and the nodes in its \textit{perturbed} neighborhood.

In our formulation, if the neighborhood remains unchanged, then $d_o(t)=d_p(t)$, and hence distortion is $0$. The distortion will be maximized if the distance of the target node to the original neighbors become significantly higher in the perturbed space than the distance to the neighbors in the perturbed graph.
With the above definition of distortion in the embedded space, the maximization objective in Eq.~\ref{eq:nphard} is approximated as : 
\begin{equation} \label{eq:new_obj}
  \CG^*=\arg \max_{\CG': d(\CG,\CG')\leq \CB}{\widehat{\delta}(t, \CG, \CG')}  
\end{equation}
This approximation is required due to two reasons: \textbf{(1)} Maximizing Eq.~\ref{eq:nphard} is NP-hard, and \textbf{(2)} Eq.~\ref{eq:nphard} is not differentiable.
\subsection{Learning $Q$-function}
\label{sec:qlearning}
We optimize Eq.~\ref{eq:new_obj} via $Q$-learning \cite{sutton2018reinforcement}, which inherently captures the combinatorial aspect of the problem in a budget-independent manner.
As depicted in Fig.~\ref{alg:dqn}, our deep Q-learning framework is an  \textit{end-to-end} architecture with a sequence of two separate neural components, a graph neural network (in experiments, we use GCN~\cite{kipf2017iclr}) and a separate Multi-layer Perceptron (MLP) with the corresponding learned parameter set $\Theta_Q$. Given a set of edges $S$ and an edge $e\not\in S$, we predict the $n$-step reward, $Q_n(S,e)$, for including $e$ to the solution set $S$ via the surrogate function $Q'_n(S,e;\Theta_Q)$.  

\textbf{Defining the framework:} The $Q$-learning task is defined in terms of state space, action space, reward, policy, and termination condition. 

$\bullet$ \textbf{State space:} The state space characterizes the state of the system at any time step $i$. Since our goal is to distort the neighborhood of target node $t$, we define it in terms of its $k$-hop neighborhood  $\mathcal{N}^k_{\CG_i}(t)$. Note that the original graph evolves at each step through edge edits. $\CG_i$ denotes the graph at time step $i$. 
The representation of state $s_i$ is defined as:
\begin{equation}
\label{eq:statespace}
    \boldsymbol{\mu}_{s_i} = {\sum_{v \in \mathcal{N}_{\CG_i}^k(t)}}{\boldsymbol{\mu}_v}
\end{equation}
where $\boldsymbol{\mu}_v$ is the embedding of node $v$ learned using a \gnn (\gcn in our implementation). 

$\bullet$ \textbf{Action: }An action, $a_i$ at $i$-th step corresponds to adding or deleting an edge $e=(v,t)$ to the solution set $S_{i-1}$ where the target node is $t$. The representation of the action $a_i$ is: 
\begin{equation}
\label{eq:action}
    \boldsymbol{\mu}_{a_i} = \pm \; \textsc{Concat} \left (\boldsymbol{\mu}_v, \boldsymbol{\mu}_t\right ),
\end{equation}
where the sign denotes edge addition ($+$) and deletion ($-$) actions.

$\bullet$ \textbf{Reward: }The immediate ($0$-step) reward of  action $a_i$ is its \textit{marginal gain} in distortion, i.e.,
\begin{equation}
    r(s_i, a_i) = \widehat{\delta}(t, \CG_{i}, \CG_{{i+1}})
\end{equation} 
where $\CG_{i+1}$ is the graph formed due to $a_i$ on $\CG_i$.

$\bullet$ \textbf{Policy and Termination: } At the $i$-th step, the policy $\pi(e\mid S_{i-1})$ selects the edge with the highest \emph{predicted} $n$-step reward, i.e., $\arg\max_{e \in C_t} Q'_n(S_i,e;\Theta_Q)$, where
\begin{equation}
\label{eq:reward}
    Q'_n(S_i,e=(u,t);\Theta_Q) = \cw_1^T\cdot \sigma(\boldsymbol{\mu_{s_i,a_i}}),\\ \text{where} \, 
\end{equation}
\begin{equation*}
\label{eq:reward2}
     \boldsymbol{\mu_{s_i,a_i}} = \cW_2\cdot\textsc{Concat} \left (\boldsymbol{\mu}_{s_i},\boldsymbol{\mu}_{a_i} \right )
\end{equation*}
$\cw_1$, $\cW_2$ are learnable weight vector and matrix respectively. We terminate training after $T$ samples.


\textbf{Learning the Parameter Set $\Theta_Q$: } 
$\Theta_Q$ consists of weight vector  $\cw_1$, and weight matrices $\cW_2$, and $\{\cW^l_{GCN}\}$. While $\cw_1$ are $\cW_2$ are used to compute $Q'_n(S_i,e)$,  $\cW^l_{GCN}$ are parameters of the \gcn for each hidden layer $l\in[1,k]$. 
$Q$-learning updates parameters in a single episode via \textit{Adam optimizer} \cite{adam} minimizing 
the \textit{squared loss}.
\vspace{-0.15in}


\begin{equation}
\nonumber
J(\Theta_Q)=(y - Q'_n(S_i, e_i; \Theta_Q))^2 \\
\text{, where }
\end{equation}
\begin{equation}
\nonumber
 y=\gamma \cdot  \max_{e=(u,t),\:u\in C}\{Q'_n(S_{i+n},e;\Theta_Q)\}+ \sum_{j=0}^{n-1}r(S_{i+j},e_{i+j})
\end{equation}
The \emph{discount factor} $\gamma$ balances the importance of immediate reward with the predicted $n$-step future reward~\cite{sutton2018reinforcement}.
The pseudocode with additional details is provided in the full version~\cite{sharma2021task}.


\subsection{Test phase}
\label{sec:test}
Given an unseen graph $\CG$ with budget $\CB$, 
 we perform $\CB$ forward passes through the \dqn. Each forward pass returns the edge with the highest predicted long-term reward and updates the state space representations. Note that the \gin is not used in the test phase. It is only used for training the \dqn. Furthermore, the proposed inference pipeline is also independent of the test \gnn or the loss function since it directly attacks the target node neighborhood. 
 

\subsection{Complexity Analysis}
Here, we analyze the running time complexity of \name during test phase. The complexity analysis for the train phase is provided in
Appendix~\ref{app:complexity} 
in the full version~\cite{sharma2021task} along with a complete proof of the obtained test-time complexity. 

Test phase involves finding $Q_n'(S_i, e)$ by doing a forward pass on a $k$-layer \gcn followed by an $L$-layer MLP. We ignore the effects of the number of layers $k$, $L$. 
Combining the costs from both \gcn and MLP, and the fact that we make $\CB$ forward passes, the total computation complexity is $O(\CB(|\CE|h_g+|\CV|h_g^2+|\CV|(h_m(1+h_g))))$. Since $h_g$ and $h_m$ may be considered as constants, the complexity with respect to the input parameters reduces to $O(\CB(\lvert\CE\rvert+\lvert\CV\rvert))$.

%% file: 5_experiment.tex
\begin{figure*}[t]
    \centering
    \vspace{-0.10in}
    \captionsetup[subfigure]{labelfont={scriptsize,bf},textfont={scriptsize}}
    \subfloat[\cora-\gcn]{\includegraphics[width=0.19\textwidth]{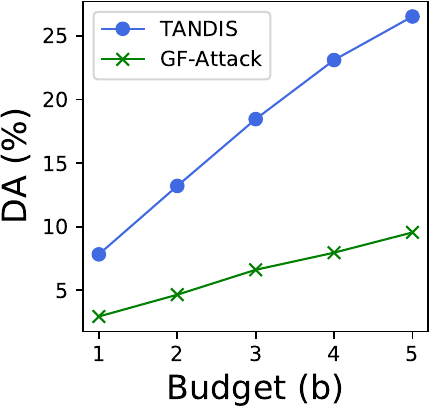}}\hfill
    \subfloat[\cora-\gsage]{\includegraphics[width=0.19\textwidth]{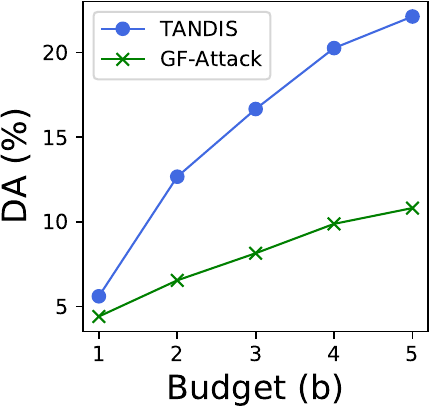}}\hfill
     \subfloat[\cora-\gat]{\includegraphics[width=0.19\textwidth]{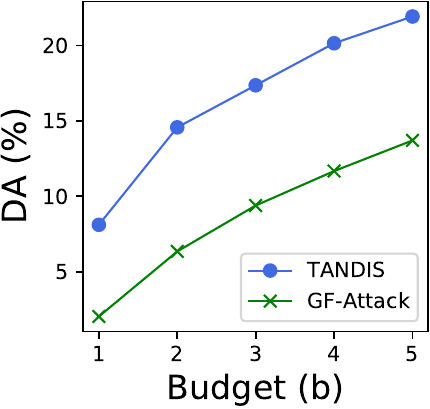}}\hfill
    \subfloat[\cora-\guard]{\includegraphics[width=0.19\textwidth]{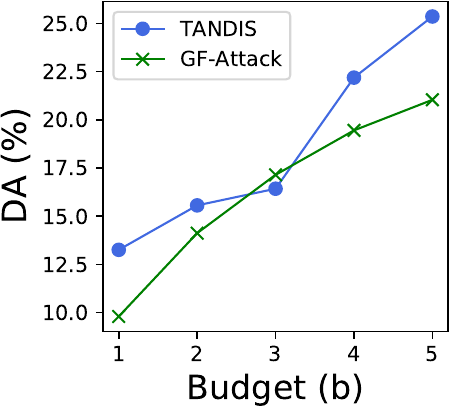}}\hfill
    \subfloat[NYC Cab][\cora-\pgnn]{\includegraphics[width=0.19\textwidth]{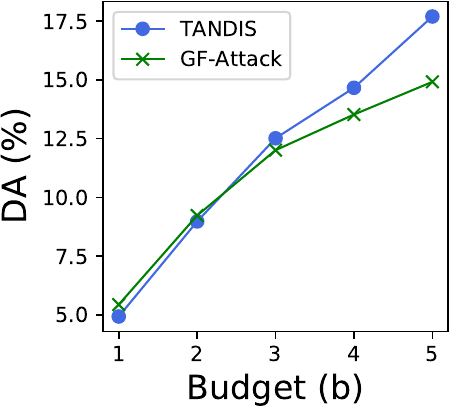}}\\
    \subfloat[\seer-\gcn]{\includegraphics[width=0.19\textwidth]{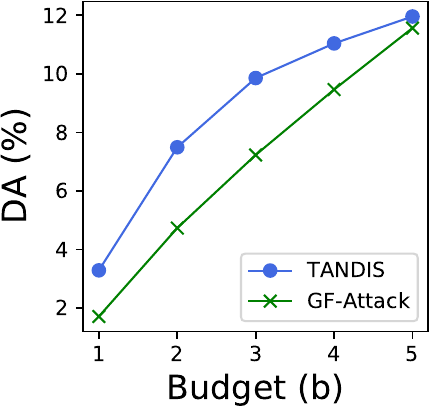}}\hfill
    \subfloat[{\scriptsize \seer-\gsage}]{\includegraphics[width=0.19\textwidth]{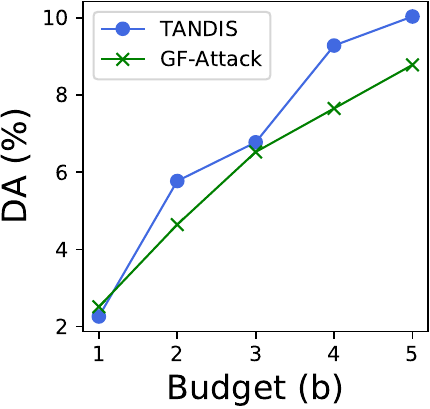}}\hfill
     \subfloat[\seer-\gat]{
     \label{fig:lpgatseer}
     \includegraphics[width=0.19\textwidth]{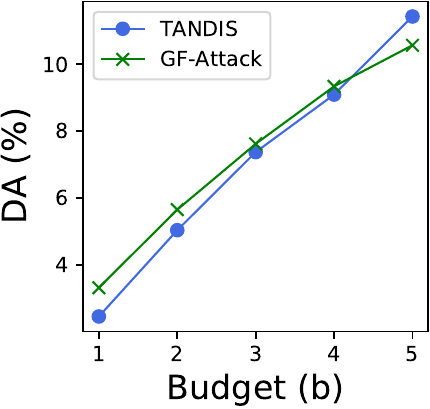}}\hfill
    \subfloat[{\scriptsize \seer-\guard}]{\includegraphics[width=0.19\textwidth]{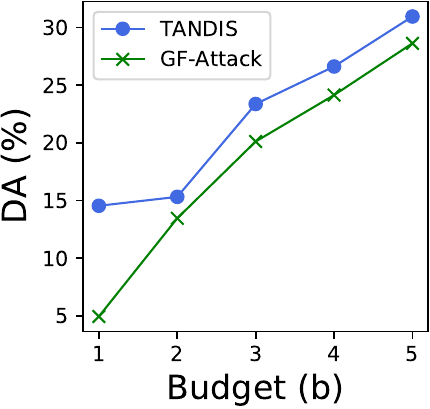}}\hfill
    \subfloat[NYC Cab][\seer-\pgnn]{
    \label{fig:lppgnnseer}
    \includegraphics[width=0.19\textwidth]{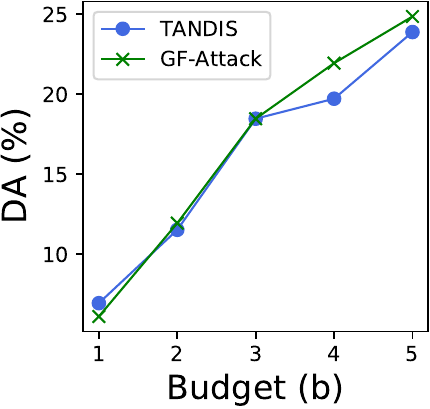}}\\
    \caption{\textbf{Link prediction: Percentage drop in accuracy (DA) in different models with varying budget in (a-d) \cora and (e-h) \seer. Higher means better.}}
    \label{fig:LP}
\end{figure*}

\begin{figure*}[t]
    \centering
    \vspace{-0.20in}
    \captionsetup[subfigure]{labelfont={scriptsize,bf},textfont={scriptsize}}
    \subfloat[\cora-\gcn]{\includegraphics[width=0.18\textwidth]{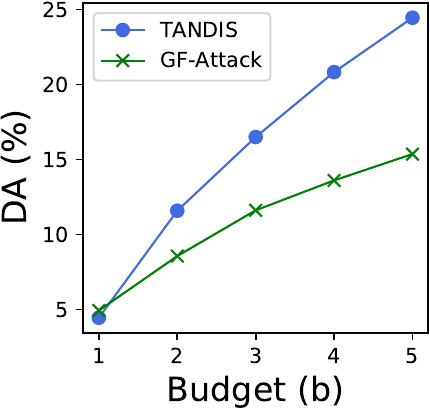}}\hfill
    \subfloat[\cora-\gsage]{\includegraphics[width=0.19\textwidth]{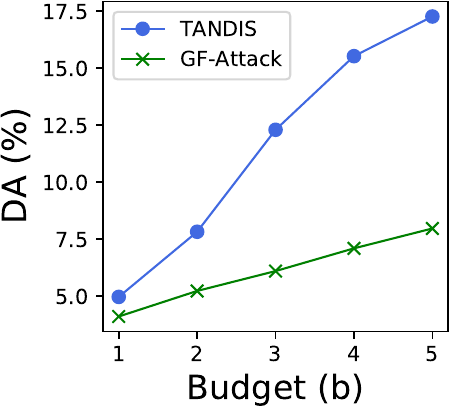}}\hfill
     \subfloat[\cora-\gat]{\includegraphics[width=0.19\textwidth]{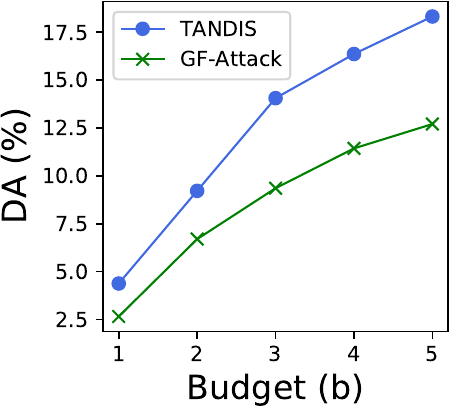}}\hfill
    \subfloat[\cora-\guard]{\includegraphics[width=0.18\textwidth]{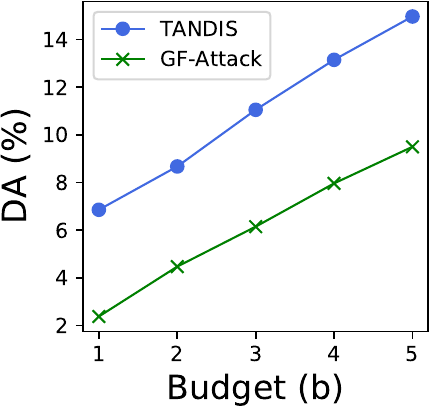}}\hfill
    \subfloat[NYC Cab][\cora-\pgnn]{\includegraphics[width=0.17\textwidth]{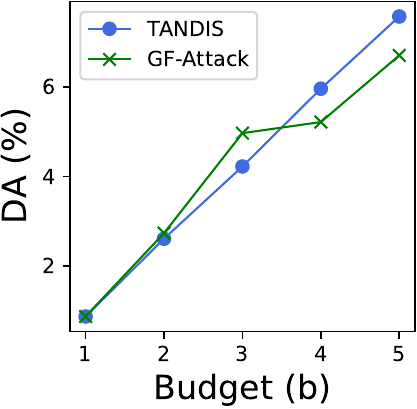}}\\
    \subfloat[\seer-\gcn]{\includegraphics[width=0.19\textwidth]{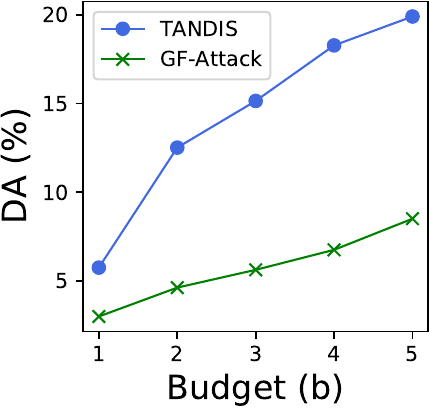}}\hfill
    \subfloat[{\scriptsize \seer-\gsage}]{\includegraphics[width=0.18\textwidth]{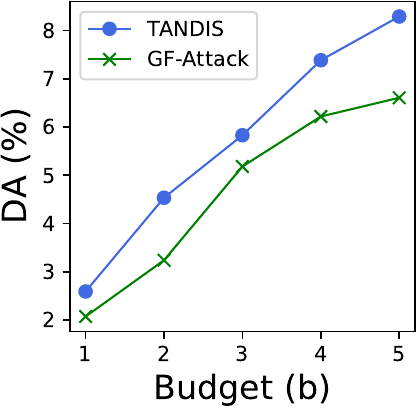}}\hfill
     \subfloat[\seer-\gat]{\includegraphics[width=0.19\textwidth]{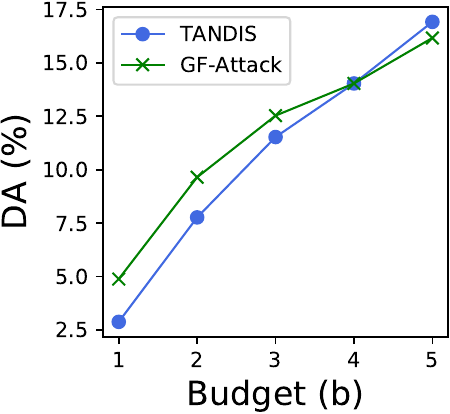}}\hfill
    \subfloat[{\scriptsize \seer-\guard}]{\includegraphics[width=0.18\textwidth]{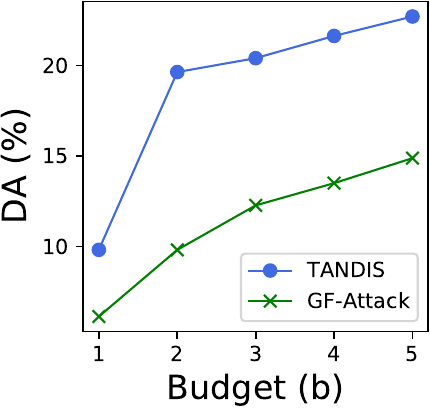}}\hfill
    \subfloat[NYC Cab][\seer-\pgnn]{\includegraphics[width=0.19\textwidth]{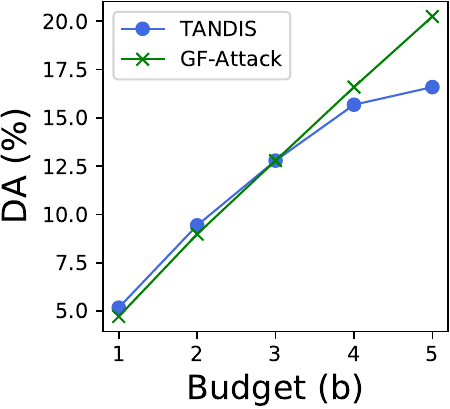}}\\
   \vspace{-0.05in}
    \caption{\textbf{Pairwise node classification: Percentage drop in accuracy (DA)  in (a-d) \cora and (e-h) \seer. Higher means better.}}
    \label{fig:pnc}
    \vspace{-0.20in}
\end{figure*}

\begin{table*}[t]
\centering
\vspace{-0.10in}
\scalebox{0.8}{
\begin{tabular}{llrrrrrrrrrrr}
    \toprule
    Budget & Dataset & \multicolumn{5}{c}{\cora} & \multicolumn{5}{c}{\seer}  \\ 
    \cmidrule(lr){3-7} \cmidrule(lr){8-12} 
     & Models & \gcn & SAGE & \gat & Guard & \pgnn & \gcn & SAGE & \gat & Guard & \pgnn \\ 
    \midrule
    (unattk) & & 81.6 & 80.8 & 83.9 & 83.4 & 78.3 & 73.9 & 76.1 & 74.9 & 75.2 & 74.6 \\
    \multicolumn{1}{c}{\multirow{3}{*}{$\CB=1$}} & \dai & 3.4 & \textbf{3.8} & 3.8 & 7.4 & \textbf{7.9} & 4.6 & 5.7 & 4.0 & \textbf{9.9} & 3.3 \\
     & \gfattack & 5.0 & 2.9 & 1.7 & 2.5 & 3.3 & 4.9 & 4.9 & \textbf{20.8} & 4.3 & 3.7 \\
     & \name & \textbf{19.2} & 2.6 & \textbf{8.0} & \textbf{12.6} & 4.9 & \textbf{33.2} & \textbf{10.8} & 6.9 & 7.5 & \textbf{4.1}  \\
     \midrule
     \multicolumn{1}{c}{\multirow{3}{*}{$\CB=5$}} & \dai & 7.6 & 7.9 & 8.8 & 15.3 & 16.6 & 10.2 & 13.8 & 11.1 & \textbf{28.6} & 8.0 \\
    & \gfattack & 22.6 & \textbf{16.8} & \textbf{21.5} & 20.6 & 15.1 & 35.5 & 22.0 & \textbf{34.7} & 18.0 & \textbf{14.8} \\
    & \name & \textbf{42.5} & 14.5 & \textbf{21.5 }& \textbf{26.7} & \textbf{29.5} & \textbf{44.9} & \textbf{23.1} & 26.4 & 13.6 & 10.7 \\
     \midrule
     \multicolumn{1}{c}{\multirow{2}{*}{$\CB=10$}} &
     \gfattack & 36.3 & \textbf{25.2} & 34.6 & 30.0 & 25.0 & 52.8 & 30.0 & 44.8 & \textbf{28.4} & \textbf{24.9} \\
    & \name & \textbf{47.2} & 23.0 & \textbf{35.2} & \textbf{31.9} & \textbf{60.0} & \textbf{54.4} & \textbf{31.1} & \textbf{44.9} & 22.7 & 19.6 \\
     \midrule
     \multicolumn{1}{c}{\multirow{2}{*}{$\CB=20$}} & 
     \gfattack & 53.6 & 38.0 & 48.3 & \textbf{42} & 45.8 & \textbf{67.5} & 38.1 & 61.4 & \textbf{39.2} & 37.8 \\
    & \name & \textbf{58.1} & \textbf{40.5} & \textbf{53.4} & \textbf{42} & \textbf{81.1} & 61.2 & \textbf{40.6} & \textbf{65.4} & 37.1 & \textbf{38.7} \\
    \bottomrule
\end{tabular}%
}
\vspace{-0.10in}
\caption{\textbf{Node Classification: Drop in DA against budget. We denote \gsage and \guard as SAGE and Guard respectively. ``unattk" represents the accuracy before attack.} }
\label{tab:nc}
\end{table*}

\section{Empirical Evaluation}
\label{sec:experiments}
We evaluate \name across \textit{three downstream tasks} of link prediction \textbf{(LP)}, node classification \textbf{(NC)} and pairwise node classification \textbf{(PNC)}~\cite{you2019pgnn,graphreach}. 
We provide additional details about the experimental setup in Appendix ~\ref{app:setup} 
of the full version~\cite{sharma2021task}.
Our code base is shared in the supplementary package.

\noindent
\textbf{Datasets:} We use three standard real-world datasets in our experiments following \cite{chang2020restricted}: \cora \cite{mccallum2000automating}, \seer and \pubmed \cite{sen2008collective}. For more details, refer
Appendix~\ref{app:setup}
of the full version~\cite{sharma2021task}.

\noindent
\textbf{Baselines:} 
We compare with \gfattack~\cite{chang2020restricted}, which is the closest in terms of features except being model-agnostic (for which, we can exploit transferability of attacks) (Recall Table~\ref{tab:summary}). We also compare with \dai~\cite{dai2018adversarial}, which is model-agnostic but not task-agnostic (limited to node classification). 
We also include non-neural baselines 
in the full version~\cite{sharma2021task} such as 
(i)~\textit{random perturbations}, (ii)~\textit{degree centrality}, and (iii)~\textit{greedy}. 

In this context, \citet{bojchevski2019adversarial} also  attacks GNNs through an idea based on neighborhood distortion. However, we do not compare against it since it is a poisoning attack. Poisoning attacks perturb the training data instead of test data as in our case. Furthermore, when this algorithm is adapted for evasion attacks, our main baseline \gfattack~\cite{chang2020restricted} outperforms significantly.
\looseness=-1

\noindent
\textbf{Target models:} 
We test the efficacy our our attacks on five state-of-the-art \gnn models in our experiments: \textbf{(1)} \gcn \cite{kipf2017iclr}, \textbf{(2)} \gsage \cite{hamilton2017graphsage}, \textbf{(3)} \gat \cite{iclr2018gat}, \textbf{(4)} \pgnn \cite{you2019pgnn}, and \textbf{(5)} \guard \cite{zhang2020gnnguard}. The first three are based on \textit{neighborhood-convolution}, \pgnn is locality-aware, and \guard is a \gnn model specifically developed to avert adversarial poisoning attacks. 

\noindent
\textbf{Parameters:} We use a $2$-layer \gin in \name to quantify distortion while training. To train the attack models, we use $10\%$ nodes each for training and validation and $80\%$ for test. On the test set,  we  ensure that the class distribution is balanced by down-sampling the majority class. Other details regarding the hyperparameters are mentioned in
~\citet{sharma2021task}.

\noindent
\textbf{Metric:} We quantify the attack performance using the percentage change in accuracy before and after attack, i.e. \emph{Drop-in-Accuracy (DA\%)}, defined as
\begin{equation}
    DA(\%)=\frac{\text{Original Accuracy}-\text{ Accuracy after attack}}{\text{Original Accuracy}}\times 100.
    \label{eq:da}
\end{equation}

\subsection{Impact of Perturbations}
In this section, we compare \name with relevant baselines on \seer and \cora. Since both \gfattack and \dai fail to scale for higher budgets on \pubmed, we defer the results on \pubmed to Section~\ref{sec:pubmed}. 

\noindent\textbf{Link Prediction (LP):} Fig.~\ref{fig:LP} evaluates the performance on LP. \name consistently produces better results i.e., higher drop in accuracy, in most cases. The only exception are \gat and \pgnn in \seer (Figures ~\ref{fig:lpgatseer}, \ref{fig:lppgnnseer}). In these cases, the drop in accuracy is similar for \name and \gfattack. Another interesting observation that emerges from this experiment is that the drop in accuracy in \guard is higher than the standard \gnn models. This is unexpected since \guard is specifically designed to avert adversarial attacks. However, it must be noted that \guard was designed to avert poisoning attacks, whereas \name launches an evasion attack. This indicates that defence mechanisms for poisoning attacks might not  transfer to evasion attacks.\\
\textbf{Pairwise Node Classification (PNC):} Here, the task is to predict if two nodes belong to the same class or not. Figure~\ref{fig:pnc} shows that \name is better across all models except in \gat on \seer, where both obtain similar performance. \\
\textbf{Node Classification (NC):} Table~\ref{tab:nc} presents the results on NC. 
Since \dai is prohibitively slow in inference, we report results till $\CB=5$ for \dai. \name  outperforms \gfattack and \dai in majority of the cases. However, the results in NC are more competitive. Specifically, \gfattack outperforms \name in a larger number of cases than in LP and PNC. A deeper analysis reveals that attacking NC is an easier task than LP and PNC. This is evident from the observed drop in accuracy in NC, which is significantly higher than in LP and PNC. Consequently, it is harder  for one technique to significantly outshine the other.

\textbf{Summary:} Overall, three key observations emerge from these experiments. First, it is indeed possible to launch task and model agnostic attacks on \gnn models. Second, \name, on average, produces a drop in accuracy that is more than $50\%$ (i.e., 1.5 times) higher than \gfattack. Third, neighborhood distortion is an effective mechanism to perturb graphs and, hence, potential solutions to defence strategies may lie in recognizing and deactivating graph links that significantly distort node neighborhoods. 

\subsection{Impact on \pubmed}
\label{sec:pubmed}
\begin{figure*}[]
    \centering
    \subfloat[NC]{
        \label{fig:pubmed_GRAND}
    \includegraphics[scale=0.5]{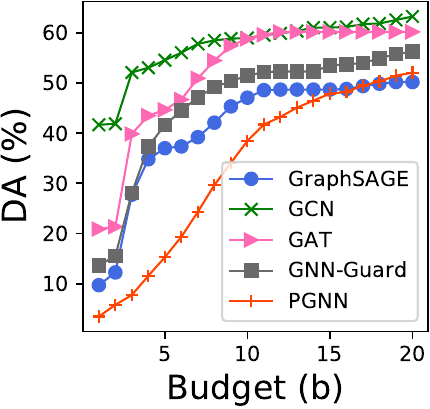}
    }
     \subfloat[LP]{
        \label{fig:pubmedLP_GRAND}
    \includegraphics[scale=0.5]{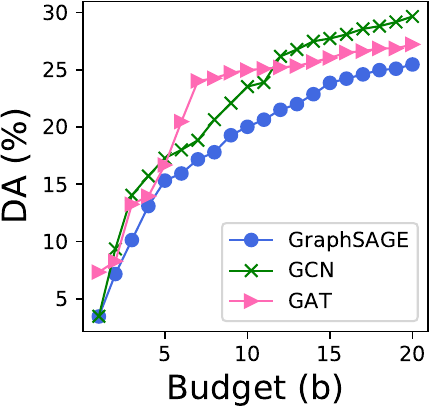}
    }
    \subfloat[PNC]{
        \label{fig:pubmedPNC_GRAND}
    \includegraphics[scale=0.5]{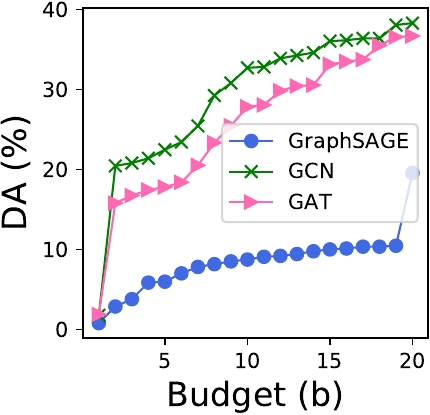}
    }
    \subfloat[Inference Time]{
        \label{fig:time}
    \includegraphics[scale=0.5]{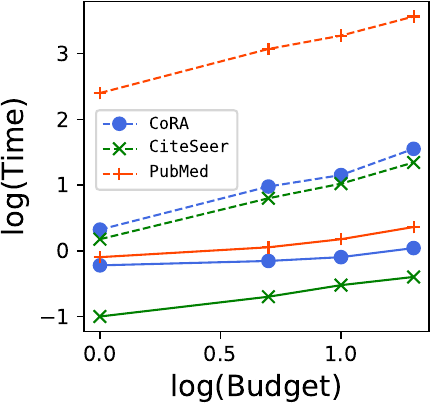}  }
    \caption{\textbf{Performance of \name in the \pubmed dataset (a-c), (d) Inference times of \name (solid lines) and \gfattack (dashed lines) against budget.}}
    \label{fig:pubmed}
\end{figure*}

\textbf{Efficacy:} Consistent with previous results, the impact of \name on NC is higher than in LP and PNC (See Figs.~\ref{fig:pubmed_GRAND}-\ref{fig:pubmedPNC_GRAND}). Furthermore, we note that \pgnn and \gsage are more robust than other models. We are unable to include \gfattack in these plots due to its high inference times. Nonetheless, we compare with \gfattack and \dai for budget $ =1$ in Table~\ref{tab:pubmed}. As visible, \name obtains the best performance in most of the cases.

\begin{table}[]
    \centering
    \subfloat[LP and PNC]{
    \scalebox{0.65}{
        \begin{tabular}{rrrrrrr}
        \toprule
        & \multicolumn{3}{c}{Link Prediction} & \multicolumn{3}{c}{Pairwise NC} \\ 
        \cmidrule(lr){2-4} \cmidrule(lr){5-7} 
        Models & \gcn & SAGE & \gat & \gcn & SAGE & \gat \\
        \midrule
        \gfattack & \textbf{5.6} & \textbf{4.4} & 6.4 & 0.2 & 0.6 & 0.4 \\
        \name & 3.5 & 3.4 & \textbf{7.4} & \textbf{0.8} & \textbf{1.8} & \textbf{1.2} \\
        \bottomrule
    \end{tabular}}}
    \hfill
    \subfloat[NC]{
    \scalebox{0.65}{
    \begin{tabular}{lrrrrrr}
    \toprule
        Models & \gcn & SAGE & \gat & \pgnn & \guard \\
        \midrule
        (unattacked) & 86.2 & 86.5 & 87.6 & 86.3 & 83.6 \\
         \dai & 2.8 & 2.3 & 2.7 & 2.1 & 5.2  \\
         \gfattack & 36.9 & 3.8 & \textbf{36.5} & 3.4 & 11.7  \\
         \name & \textbf{41.6} & \textbf{9.7} & 20.9 & \textbf{3.5} & \textbf{13.7} \\ 
        \bottomrule
    \end{tabular}
    }}
    \vspace{-0.10in}
    \captionof{table}{\textbf{Drop in accuracy (DA) in \pubmed at budget $=1$.} }
     \vspace{-0.10in}
    \label{tab:pubmed}
\end{table}

\subsection{Running Time}
Figure ~\ref{fig:time} compares the inference times of \name and the second-best baseline \gfattack, i.e. the running time of finding black-box perturbations. We do not compare with \dai, since it fails to complete running for any $\CB>1$ even after $10$ hours. \name (represented by solid lines) is more than $3$ orders of magnitude faster than \gfattack for the largest dataset, \pubmed. Also, time taken by \name grows at a much lower rate with the increase in budget than \gfattack. Furthermore, \name requires training only once per dataset and the entire training procedure takes upto $3$ hours even for the largest dataset, i.e. \pubmed. 

\subsection{Potential Detection Mechanisms}
\label{sec:detection}

\begin{table}[b]
    \centering
    \scalebox{0.9}{
    \begin{tabular}{cc}
        \toprule
        Feature & Correlation (p-value) \\
        \midrule
        Feature Similarity (fs) & $\mathbf{- 0.20^* \pm 0.196}$  \\
        Degree (dg) & $\mathbf{- 0.15^* \pm 0.008}$ \\
        Local Clustering Coeff. (cc) & $\mathbf{+ 0.25^* \pm 0.004}$ \\
        \bottomrule
    \end{tabular}}
    \caption{\textbf{Correlation of node properties with neighborhood distortion (${}^*$ indicates $p < 0.001$).}}
    \label{tab:interpretability}
\end{table}

Here, we conduct an experiment on \cora dataset to find correlations between perturbations chosen by our attacks and the ones chosen according to common network properties. The aim is to \textbf{(1)} interpret our attacks in graph space and \textbf{(2)} identify potential ways to catch our evading attack. 

In particular, we consider all node pairs of the form $(t,v)$, where $t$ is the target and $v$ is the node to which an edge may be added or removed. We characterize each $v$ with several graph level properties such as clustering coefficient, conductance, etc (See
\citet{sharma2021task}
for the complete list). Corresponding to each property, we create a sorted list $P$ of nodes in descending order of their property values. Similarly, we create another list $D$ with nodes that are sorted based on the distortion produced due to removal/addition of edge $(t,v)$. We next compute the \textit{Spearman's rank correlation} coefficient between $P$ and $D$ and its $p$-value. Properties with statistically significant correlations may indicate graph-level features that govern the selection of perturbations by \name and thus, effective in detecting these attacks. 

Table~\ref{tab:interpretability} presents the properties that are statistically significant. 
These results indicate that edges to nodes of high \textit{clustering coefficient (cc)} lead to large distortion. This result is not surprising since a node with high cc short-circuits the target node to a large number of nodes (or removes this short-circuit in case of edge deletion). The negative correlation with \textit{fs} is also intuitive, since low \textit{fs} with a node indicates that an edge to it injects a large amount of heterophily. The results for \textit{dg} is rather surprising, where a negative correlation indicates that edges to low-degree nodes are more effective. A deeper analysis reveals that \textit{dg} is negatively correlated with \textit{cc}, which may explain the negative correlation with distortion as well. Note that, in \gfattack, it has been shown that perturbing edges with high degree nodes is not an effective attack strategy. Our study is consistent with that result, while also indicating that \textit{cc} is a more effective graph-level parameter.

%% file: 7_conclusion.tex
\section{Conclusions}
\label{sec:conclusions}

\gnns have witnessed widespread usage for several tasks such as node classification and  
link prediction.
Hence, assessing its robustness to practical adversarial attacks is important to test their applicability in security-critical domains. While the literature on adversarial attacks for \gnns is rich, they are built under the assumption that the attacker has knowledge of the specific \gnn model used to train and/or the prediction task being attacked. The key insight in our proposed work is that simply distorting the neighborhood of the target node leads to an effective attack regardless of the underlying \gnn model or the prediction task. 
Hence, hiding the model or the task information from the attacker is not enough.
We also find that such perturbations are not correlated with simple network properties and may be hard to detect. 
Our work thus opens new avenues of research that can focus on defending and detecting such practical attacks to analyze why such perturbations should transfer across different tasks and victim models. 

%% file: 8_appendix.tex
\appendix
\section*{Appendix}
\renewcommand{\thesubsection}{\Alph{subsection}}

\subsection{Attack taxonomy}
\label{app:taxonomy}

\begin{itemize}
\item \textbf{Poisoning attacks:} In poisoning attacks, the attacker injects adversarial samples in the training data, so that the accuracy of the predictive task being modeled is compromised \cite{bojchevski2019adversarial,gupta2021adversarial,li2020adversarial,liu2019unified,sun2020www,wang2019acm,xu2019ijcai,zhang2019ijcai,zugner2018adversarial,zugner_adversarial_2019}. 
Note that \cite{bojchevski2019adversarial} also addresses attacks on embedding, however we do not compare it against our method as it is a poisoning attack and one of our baselines, \gfattack~\cite{chang2020restricted} clearly outperforms this on the ode classification task.

\item \textbf{Evasion attacks:} In evasion attacks, the attacker adversarially modifies the test graph, so that the accuracy on the test graph is compromised \cite{chang2020restricted,chen2021graphattacker,dai2018adversarial,ma2020towards,ma2019attacking,wang2020efficient,wu2019ijcai,xu2019ijcai,zugner2018adversarial}. A modification may involve addition or deletion of nodes and edges, or changing node attributes. Note that the trained model remains unchanged.
\end{itemize}

Since an evasion attack relies only on the test data, it is  comparatively easier to launch than a poisoning attack. As an example, consider the task of friend recommendation (link prediction) in a social network such as Facebook. It is easier for an attacker to infuse bots in Facebook that send friend requests (edge rewiring) to strategically selected users than getting access to Facebook's training model.

\subsection{\textsc{Proof of Thm.~\ref{thm:nphard}}}
\label{app:nphard}
\subsubsection{NP-hardness}
We reduce the \SC problem to our maximization objective for $k=2$. 
\begin{definition}[Set Cover]
Given a collection of subsets $S=\{S_{1},\cdots,S_{m}\}$ from a universe of items $U=\{u_{1},\cdots,u_{n}\}$ and a budget $b$, determine if there exists a set of $b$ subsets $\mathbb{\mathcal{A}}\subseteq S$ such that  $\cup_{\forall S_i\in \mathbb{\mathcal{A}}} S_i= U,\;|\mathcal{A}|=b$.
\end{definition}
Let $\langle U=\{ u_{1},\cdots,u_{n} \},S=\{S_{1},\cdots,S_{m}\}, b\rangle$ be an instance of the \SC problem. We map this to an instance of our problem via constructing a graph $\CG=(\CV,\CE)$ as follows. 
From $S$ and $U$, we create two sets of nodes $X=\{x_i\mid S_i\in S\}$ and $Y=\{y_j\mid u_j\in U\}$ respectively. We also add an additional node $t$ with no edges attached to it. Thus, $\CV=X \cup Y\cup\{t\}$. Additionally, if $u_j \in S_i$, an edge $(x_i,y_j)$ is added to the edge set $\CE$. The set of colluding nodes is set to $\CC=X$, budget $\CB=r$, and the target node is $t$. In this construction, if $\mathcal{A}$ is a solution set of the \SC instance, then the corresponding set of edge additions $B = \{(t,x_i)|S_i\in \mathcal{A}\}$ maximizes $\delta(t,\CG,\CG')$. Specifically, $\CN^2_{\CG}(t)=\emptyset$ since no edges are incident on it. Thus, maximizing $\delta(t,\CG,\CG')$ is equivalent to maximizing $|\mathcal{N}^2_{\CG'}(t)|$. The maximum value of  $|\CN^2_{\CG'}(t)|=b+n$ if and only if $\mathcal{A}$ is a set cover. $\hfill\square$

\subsubsection{Non-montone}
We first define $\delta(v,\CG,\CG')$ as a set function:

\begin{equation*}
    \delta(v,\CG,\CG') := f\left(\mathcal{N}^k_{\CG'}(v); \mathcal{N}^k_{\CG}(v)\right) = 1 - \frac{\left|\mathcal{N}^k_{\CG}(v) \cap \mathcal{N}^k_{\CG'}(v) \right|}{\left|\mathcal{N}^k_{\CG}(v) \cup \mathcal{N}^k_{\CG'}(v) \right|}
\end{equation*}

Hence, the objective can be written as a set function as

\begin{equation*}
    f(A;B) = 1 - \frac{|A \cap B|}{|A \cup B|}
\end{equation*}

Consider a set $S$ and $x \not\in S$. Now, we have two cases:

\paragraph{Case (1) $x \in B$:} $f(S \cup \{x\}; B) - f(S; B) = \frac{|S \cap B|}{|S \cup B|} - \frac{|(S \cup \{x\}) \cap B|}{|(S \cup \{x\}) \cup B|} = \frac{-1}{|S \cup B|} < 0.$

\paragraph{Case (2) $x \not\in B$:} $f(S \cup \{x\}; B) - f(S; B) = \frac{|S \cap B|}{|S \cup B|} - \frac{|(S \cup \{x\}) \cap B|}{|(S \cup \{x\}) \cup B|} = |S \cap B|\left(\frac{1}{|S \cup B|} - \frac{1}{|S \cup B| + 1}\right) > 0.$

Hence, $f$ is not monotonic and the sign of its change depends on whether $x \in B$ or $x \not\in B$. $\hfill\square$

\subsubsection{Non-submodular}

\begin{definition}[Submodularity]
    A set-function $h$ is submodular if for every sets $X, Y$ such that $X \subseteq Y$, and for every $e \not\in Y$, 
    \begin{equation*}
        h(X \cup \{e\}) - h(X) \geq h(Y \cup \{e\}) - h(Y).
    \end{equation*}
\end{definition}

Consider $f$ as defined above. We consider two sets $S, T$ such that $S \subset T$. Let $x \not\in T$ but $x \in B$. Then, $f(S \cup \{x\}; B) - f(S;B) = \frac{-1}{|S \cup B|}$, and $f(T \cup \{x\}; B) - f(T;B) = \frac{-1}{|T \cup B|}$. Thus, $(f(S \cup \{x\}; B) - f(S;B)) - (f(T \cup \{x\}; B) - f(T;B)) = \frac{1}{|T \cup B|} - \frac{1}{|S \cup B|} < 0$. 

Therefore, the objective $f$ is not submodular. $\hfill\square$

\subsection{Expressive power of \gin}
\label{app:gin}


\subsubsection{Pseudo-code of \gin}

In Section \ref{sec:train_gen}, \gin is trained in an unsupervised manner minimizing the loss given in Equation \ref{eq:loss}. In particular, we apply negative-sampling \cite{hamilton2017graphsage} to approximate the following loss function:
\begin{equation}\small
    \CL(\SM(\CG)) = -\log(\sigma(\CZ_u^T \CZ_v)) - \ \mathbb{E}_{v_n \sim P_n(v)} \log (\sigma(-\CZ_u^T \CZ_{v_n}))
    \label{eq:negsamp}
\end{equation}

where $P_n(v)$ denotes a distribution of nodes which do not have an edge with $v$ and $v_n$ is negatively sampled from this distribution. $\sigma$ denotes the activation function Sigmoid. 

For training, we sample a set of positive and negative random walks and optimize the total loss in a batch for a fixed number of epochs. Therefore, $v$ in Equation \ref{eq:negsamp} denotes the nodes in a positive random walk, which is derived from a random walker. Similarly, $v_n$ denotes the nodes in a negative random walk, i.e., $v_n$ is a randomly chosen node in the whole graph. Algorithm \ref{alg:gin} shows the entire process. 

\begin{algorithm}[h]
\caption{Training the \gin model $\CM_G$}
\label{alg:gin}
\begin{algorithmic}[1] 
\REQUIRE Number of epochs $N_e$, hyperparameters $h_{RW}$ (See App. \ref{app:gin_param}) for the Random Walker
\ENSURE Learn parameter set $\Theta_G$
\FOR{epoch $i \leftarrow 1$ to $N_e$}
\STATE $W_p \leftarrow $ Sample a batch of positive random walks with $h_{RW}$
\STATE $W_n \leftarrow $ Sample a batch negative random walks by selecting a set of random nodes

\STATE Back-propagate the average loss (Equation \ref{eq:negsamp}) using $\CM_G$ on $W_p$ and $W_n$
\ENDFOR
\RETURN $\Theta_Q$
\end{algorithmic}
\end{algorithm}

\subsubsection{Expressive Power of \gin}
In this section, we explain the reasons of why a \gin model would be the most effective choice in the case of a black-box model. Since our attack must be model-agnostic, the embedding neighborhood should be such that it can represent any underlying model's representation. \gin has been shown to be a maximally powerful \gnn model. In particular, it has been shown to be as powerful as a \textit{Weisfeiler-Lehman (WL) graph isomorphism test}, which assigns canonical labels such that unique labels guarantee that the two graphs are not isomorphic. \cite{xu2019gin} shows that injectivity of the neighborhood aggregation function is a necessary condition for a \gnn to simulate a WL test and give one such example in \gin. Hence, any underlying black-box \gnn model has less representation power by aggregating information from their neighbors than a \gin. This motivates us to use a \gin to model the neighborhood for distortion, so that we can effectively attack any target black-box model. 

\begin{figure*}[t]
    \centering
    \captionsetup[subfigure]{labelfont={scriptsize,bf},textfont={scriptsize}}
    \subfloat[\cora-\gcn]{\includegraphics[width=0.19\textwidth]{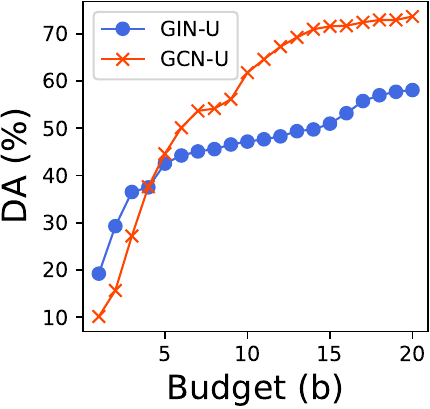}}\hfill
    \subfloat[\cora-\gsage]{\includegraphics[width=0.19\textwidth]{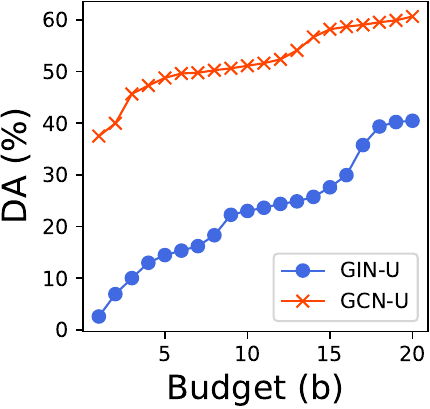}}\hfill
     \subfloat[\cora-\gat]{\includegraphics[width=0.19\textwidth]{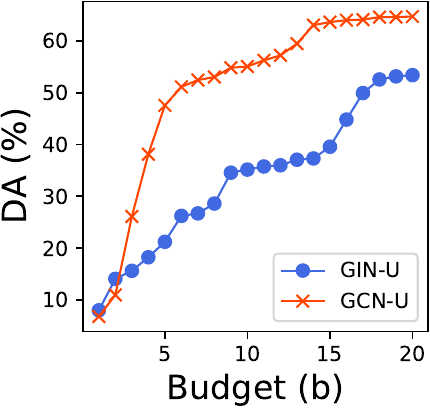}}\hfill
    \subfloat[\cora-\guard]{\includegraphics[width=0.19\textwidth]{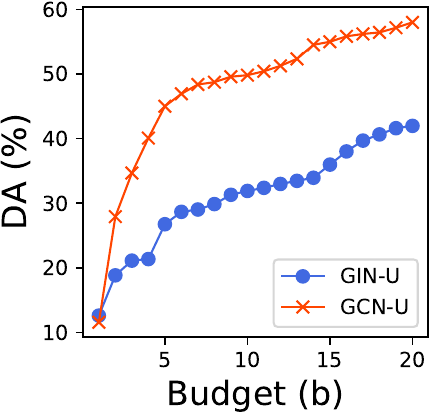}}\hfill
    \subfloat[\cora-\pgnn]{\includegraphics[width=0.19\textwidth]{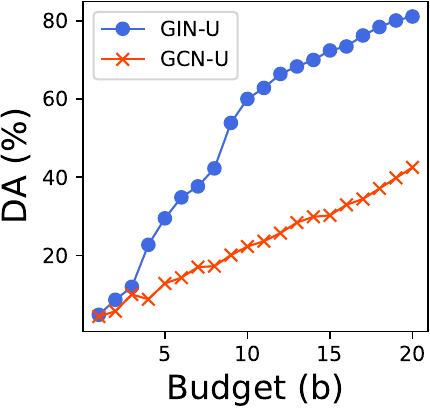}\label{fig:pgnn_gcn_gin}}
    \caption{\textbf{Ablation study: Comparison of using \gcn v/s \gin embeddings for distortion in the \name attack for the node classification task in \cora. \gin-U represents that \gin is trained in an unsupervised manner.}}
    \label{fig:abl}
\end{figure*}

In order to further justify our claim, we conduct an ablation study and compare the performance of our attacks using a \gcn and a \gin as our embedding model. Both are trained in the same unsupervised manner. Figure \ref{fig:abl} compares the $DA(\%)$ across different budgets when using \gcn embeddings as compared to when using \gin embeddings for node classification in \cora dataset. The results are interesting and show the apparent trade-off in using a theoretically powerful \gin model to attack common \gnn target models. While \gcn embeddings are able to transfer the neighborhood better (our method produces more DA, higher is better) for \gnns that are based on \textit{neighborhood convolution}, they do not get transferred well to the \textit{locality-aware} \pgnn (Figure \ref{fig:pgnn_gcn_gin}).

Hence, this shows that our strategy can be tuned based on the amount of information that is made available to the attacker. In a completely model-agnostic setting, it is reasonable to use a powerful \gin model to generate embeddings. If we know some distinct features about the underlying model, then it will be more beneficial to use an embedding model that has a similar or most representative neighborhood aggregation function, so that the knowledge of neighborhood distortion gets transferred well. 

Furthermore, it is worthwhile to note here that if the attacker wishes to leverage the task-agnosticism, then they might even use an embedding model trained for that specific task (in which case, the embeddings/representations would be specially tuned to the attack scenario). \name is a generic and unique framework that allows this kind of tuning according to the available attack/model information.

\subsection{Rationale behind two GNNs}
\label{app:ratio_gnn}

The \gnns serve different roles. The role of $\boldsymbol{\mu}_v$ is to effectively characterize the state-space (Eq.~\ref{eq:statespace}) and actions (Eq.~\ref{eq:action}), and capture the combinatorial nature of the problem. Hence, the \gcn embeddings $\boldsymbol{\mu}_v$ are trained end-to-end in the \dqn alongwith  an MLP to learn the $Q$-function. On the other hand, \gin embeddings $\CZ_v$ are trained independently to characterize neighborhood distortion. Furthermore, while the \gin is trained in a transductive manner with one-hot encodings of nodes, the \gcn trains in an inductive manner, where in the initial layer the nodes are characterized by their feature vectors $\CX_v$ (Recall \S~\ref{sec:formulation}).

\subsection{Greedy Selection of Edge Perturbations for Training}
\label{app:greedy}

Algorithm \ref{alg:dqn} describes how the Q-function is learned in the RL setup (Section \ref{sec:qlearning}) of \name. In particular, for each node in the training set, we generate perturbations for $T$ steps following a $\epsilon$-greedy strategy, as noted in Step \ref{alg_step:epsg} (Algorithm \ref{alg:dqn}). Each such perturbation can be represented by a tuple storing the current state, new state, action, and accumulated rewards (defined in Section \ref{sec:qlearning}). This is stored in an ``experience replay" memory, from which a batch of samples is randomly sampled. Finally, this batch is used to backpropagate the squared loss. We repeat this process for $L$ steps. 
\begin{algorithm}[h]
\caption{Learning $Q$-function}
\label{alg:dqn}
\begin{algorithmic}[1] 
\REQUIRE  Trained \gin model $\CM$, hyper-parameters $M$, $N$ relayed to fitted $Q$-learning, number of episodes $L$ and sample size $T$.
\ENSURE Learn parameter set $\Theta_Q$
\STATE Initialize experience replay memory $M$ to capacity $N$
\FOR{episode $e \leftarrow 1$ to $L$}
\FOR{step $j \leftarrow 1$ to $T$}
\STATE $v_j \leftarrow
\begin{cases}
\text{random node }v \not\in S_j; \text{ with probability }\\ \hspace{32mm}\epsilon=\max\{0.05,0.9^j\} \\
\text{argmax}_{v \not\in S_j} Q'_n(S_j,v;\Theta_Q); \text{ otherwise}
\end{cases}$ \label{alg_step:epsg}
\STATE $S_{j+1}\leftarrow S_j\cup\{v_j\}$
\IF{ $t \geq n$}
\STATE Add tuple $(S_{j-n}, v_{j-n}, \sum_{i=j-n}^{t}r(S_i,v_i), S_j)$ to $M$ 
\STATE Sample random batch $B$ from $ M$
\STATE Update $\Theta_Q$ by Adam optimizer for $B$
\ENDIF
\ENDFOR
\ENDFOR
\RETURN $\Theta_Q$
\end{algorithmic}
\end{algorithm}

\subsection{Complexity Analysis}
\label{app:complexity}
\subsubsection{Train phase}
The training complexity consists of two parts -- (1) Training the \gin, and (2) Training the \dqn. \gin is trained in an unsupervised manner, following Algorithm 1. This accounts for a training complexity of $O(N_elh_gT_r)$, where $N_e$ is the no. of epochs, $l$ is the no. of \gin layers, $h_g$ is the hidden dimension of \gin, and $T_r$ is the time for random walk generation. We follow node2vec’s code \cite{grover2016node2vec} 
to generate random walk, which accounts for $O(L_r/(k_r(L_r - k_r)))$, where $L_r$ is the length of the random walks, $k_r$ is the no. of samples to generate. 
The second part in the analysis is due to \dqn, which is trained by following Algorithm 2. This accounts for $O(LT |\CV|(Bh_mh_{gc} + (|\CE|h_{gc} + |\CV|h_{gc}^2))$, where $L$ is the no. of episodes, $T$ no. of time steps, $B$ batch size, $|\CV|$ no. of nodes, $|\CE|$ no. of edges, $h_m$ dimension of MLP, and $h_{gc}$ dimension of GCN layers respectively. $h_m$ $h_{gc}$ is the time for MLP forward pass, and $(|\CE|h_{gc} + |\CV|h_{gc}^2)$ is the time for GCN forward pass (as derived in the next section).

\subsubsection{Test phase}
To find $Q_n'(S_i, e)$, we do a forward pass on a $k$-layer \gcn followed by  an $L$-layer MLP. We ignore the factors of $k$ and $L$ as their values are usually small and fixed ($\leq 3$), and assume the average degree in $\CG$ to be $d$.

\textbf{\gcn:} We assume the hidden dimension to be $h_g$. Each node draws messages from their neighbors and performs a mean pool, followed by linear transformation through weight matrices $\cW^l_{\gcn}$. These operations consume $O(d\cdot h_g)$ and $O(h_g^2)$ respectively. When performed over all nodes in $\CG$, the complexity is $O(|\CE|h_g+|\CV|h_g^2)$.

\textbf{MLP:} The transformations performed by the MLP is outlined in Eq.~\ref{eq:reward}. The representation dimensionality of a state and an action are $h_g$ (Eq.~\ref{eq:statespace}) and $2h_g$ (Eq.~\ref{eq:action}) respectively. If the hidden dimension in the MLP is $h_m$, then computing  $\boldsymbol{\mu}_{s_i,a_i}$ in Eq.~\ref{eq:reward} consumes $O(h_g\cdot h_m)$. The final conversion to a scalar (predicted long-term rewards) consumes an additional $O(h_m)$ cost. This operation is repeated over each candidate edge since we choose the one with the highest predicted reward. Hence, the total complexity is $O(|\CV|(h_m(1+h_g)))$.

Combining the costs from both \gcn and MLP, and the fact that we make $\CB$ forward passes, the total computation complexity is $O(\CB(|\CE|h_g+|\CV|h_g^2+|\CV|(h_m(1+h_g))))$. Since $h_g$ and $h_m$ are fixed constants, the complexity with respect to the input parameters reduces to $O(\CB(\lvert\CE\rvert+\lvert\CV\rvert))$.

\subsection{Experimental Setup}
\label{app:setup}
All experiments are performed on a machine running Intel Xeon G-$6148$ with a setting of $20$ cores of $2.4$ GHz, one Nvidia $V100$ card with $32$ GB GPU memory and $192$ GB RAM with Ubuntu 16.04.

\subsubsection{Datasets}
We use three real-world datasets in our experiments following \cite{chang2020restricted}: \cora \cite{mccallum2000automating}, \seer and \pubmed \cite{sen2008collective}. These are citation networks where vertices denote documents with bag-of-words features and edges denote citation links. We consider only the largest connected component which is consistent with our baseline setup \cite{chang2020restricted}.
The statistics of the datasets are shown in the Table \ref{tab:data_stat}.

\begin{table}[t]
\centering
\begin{tabular}{lrrrr}
		\toprule
		\textbf{Datasets} & \textbf{\# Nodes} & \textbf{\# Edges} & \textbf{\# Node Labels} \\
		\midrule
		Cora & 2.7K & 5.4K & 7 \\
		Citeseer & 3.3K & 4.7K & 6 \\
		PubMed & 19K & 44K & 3\\
		\bottomrule	
	\end{tabular}
	\caption{\textbf{Datasets used in the experiments.}}
	\label{tab:data_stat}
\end{table}


\subsubsection{Tasks}
\paragraph{Node Classification:} All the target models are trained for the node classification task using $10\%$ of the nodes each for training and validation. The remaining $80\%$ nodes are used for the testing where each node is separately considered as a target node for the attack strategies and the average accuracy quantified as $DA(\%)$ in Equation \ref{eq:da}. For the large dataset Pubmed, we consider a randomly sampled $10\%$ of the nodes out of the test set and we present the average value after $5$ different runs. 

\paragraph{Link Prediction:} All the target models are trained using a binary cross entropy loss for this task using $10\%$ of the positive and negative links each for training and validation. Note that we ensure down-sampling of the majority class and thus, ensure that equal number of positive (existing) and negative (non-existing) links are considered in our training, validation and test sets. We consider a randomly sampled $30\%$ links for the test set and note the average drop in accuracy after considering each links separately. Each target link is attacked by considering one of the end-nodes as the target node on which edges can be added or deleted. We randomly sample $10\%$ of the test set for the large dataset Pubmed and note the average value after five different runs. 

\paragraph{Pairwise node-classification:} The setup is similar to the task of link prediction except for the end-labels which denote the equality of node pairs' labels rather than an existence of an edge between them. Thus, the class distribution is now balanced based on this label. We follow a $10-10-30$ split here as well and take $10\%$ sample to target Pubmed.

\subsubsection{Parameters}
\label{app:gin_param}

Our framework (Figure \ref{fig:flow_diag}) needs two trainable models, \gin and \dqn. These are trained separately and have different hyper-parameters as described below.

\paragraph{\gin:} We use a $2$-layer \gin model with hidden dimension of $16$. Here also, we use $10\%$ each for training and validation. This is trained in an unsupervised manner. To generate the random walks, we use $p=q=1$ (where $p$ and $q$ denote the likelihood of return and measure of in-out search respectively \cite{grover2016node2vec}), a walk length of $20$, a context size of $10$ and do $10$ walks per node. 

\paragraph{\dqn:} To train the \dqn component, we use Deep Q-learning as noted in the Algorithm \ref{alg:dqn}. Here, the number of episodes $L$ and number of steps $T$ are fixed to be $100$ and $10$ respectively. Also, $n$ is fixed to be equal to $2$ and thus, we consider a $2$-step Q-function. To model the \dqn, we consider a $2$-layer \gcn followed by a $2$-layer MLP, both with hidden dimensions of $16$. We use $40\%$ nodes as targets for the training the \dqn out of which the replay memory is made, following $\epsilon$-greedy steps.

\subsubsection{Baselines}

\paragraph{\dai:} We use the \dai \cite{dai2018adversarial} in a model-agnostic manner following the formulation mentioned in \gfattack \cite{wang2020efficient}. In particular, we train the Deep Q-network using a surrogate \gcn model and then transfer these attacks to all our target models (as mentioned in Section \ref{sec:experiments}). Also, note that the surrogate \gcn is trained differently than the target \gcn model. We train for $100000$ steps for each dataset and separately for each budget (until $5$ as noted in Table \ref{tab:nc}). 

\paragraph{\gfattack}\cite{wang2020efficient}: This is the state-of-the-art in the restricted black-box setting and the only available task-agnostic attack framework. It attacks the embeddings by perturbing the corresponding graph filters. While the exact optimization may depend upon the type of black-box model to attack, we can exploit the transferability of these attacks and make it model-agnostic. We use the \gcn setting mentioned in \cite{wang2020efficient} to attack each of our black-box models. For link tasks, we use the same strategy as ours and target a randomly chosen end-node.

\subsection{Additional Results}
\label{app:results}

\begin{figure*}[]
    \centering
    \captionsetup[subfigure]{labelfont={scriptsize,bf},textfont={scriptsize}}
    \subfloat[\cora-\gcn]{\includegraphics[width=0.19\textwidth]{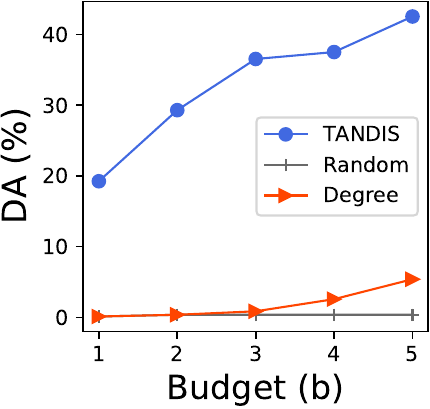}}\hfill
    \subfloat[\cora-\gsage]{\includegraphics[width=0.19\textwidth]{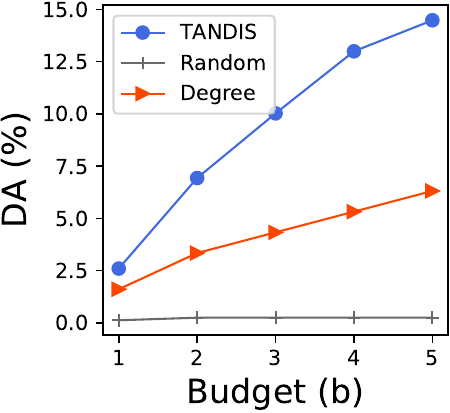}}\hfill
     \subfloat[\cora-\gat]{\includegraphics[width=0.19\textwidth]{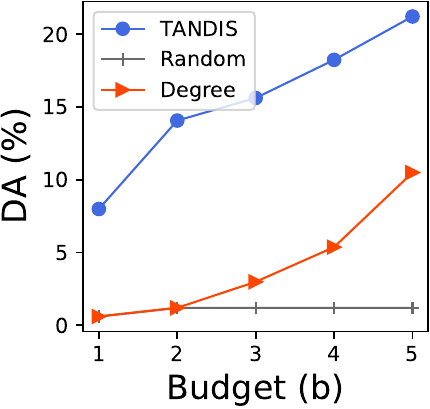}}\hfill
    \subfloat[\cora-\guard]{\includegraphics[width=0.19\textwidth]{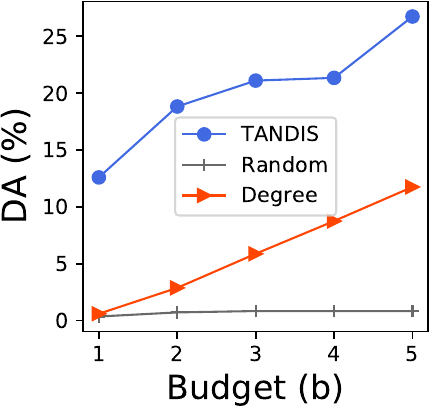}}\hfill
    \subfloat[\seer-\pgnn]{\includegraphics[width=0.19\textwidth]{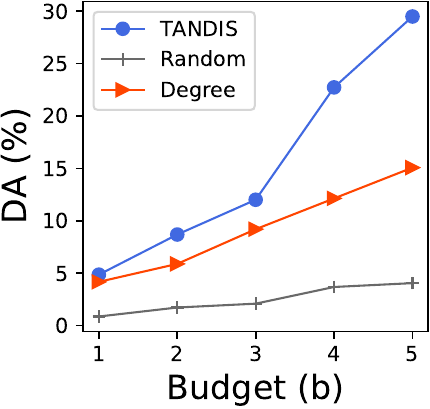}} \\
    \subfloat[\seer-\gcn]{\includegraphics[width=0.19\textwidth]{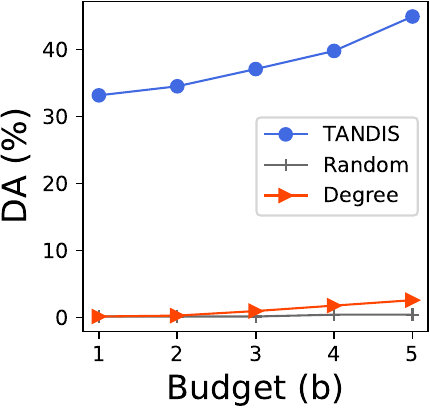}}\hfill
    \subfloat[{\scriptsize \seer-\gsage}]{\includegraphics[width=0.19\textwidth]{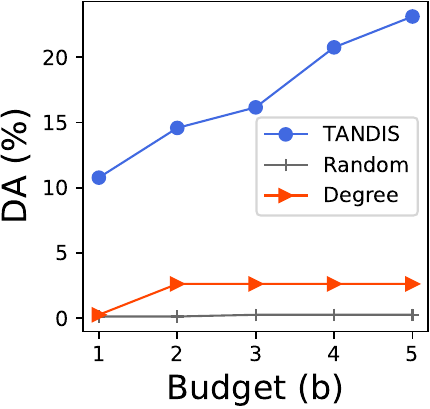}}\hfill
     \subfloat[\seer-\gat]{\includegraphics[width=0.19\textwidth]{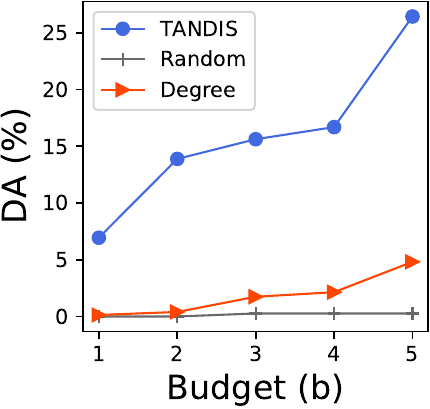}}\hfill
    \subfloat[{\scriptsize \seer-\guard}]{\includegraphics[width=0.19\textwidth]{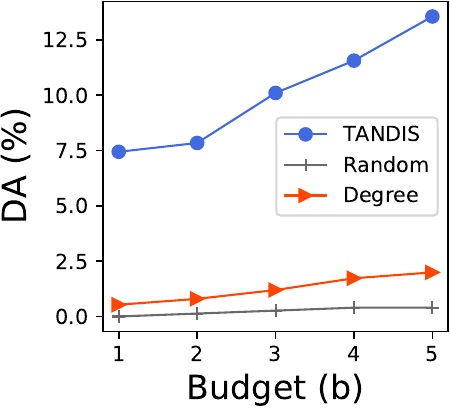}}\hfill
    \subfloat[\seer-\pgnn]{\includegraphics[width=0.19\textwidth]{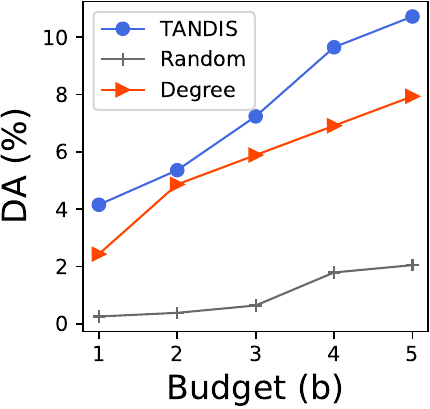}}
    \caption{\textbf{Comparison of non-neural baselines with \name for the node classification task in \cora and \seer. \name outperforms the baselines in all cases.}}
    \label{fig:non_neu_base}
\end{figure*}

\begin{figure*}[]
    \centering
    \subfloat[\cora-\gcn]{\includegraphics[width=0.23\textwidth]{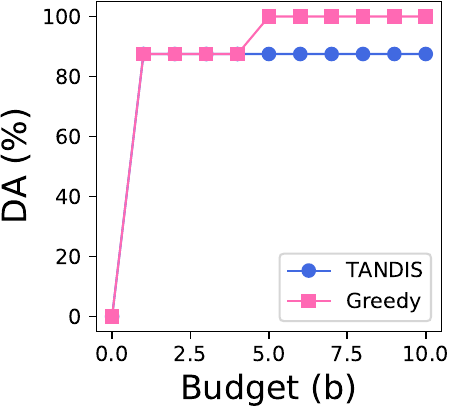}}\hfill
    \subfloat[\cora-\gsage]{\includegraphics[width=0.23\textwidth]{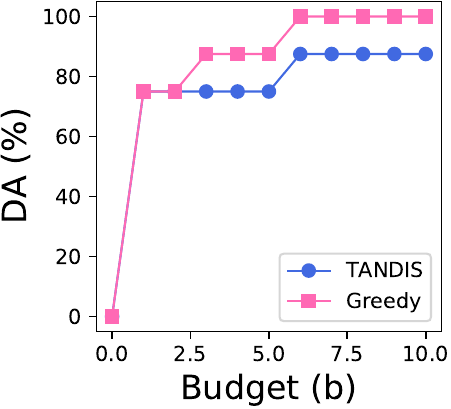}}\hfill
    \subfloat[\cora-\gat]{\includegraphics[width=0.23\textwidth]{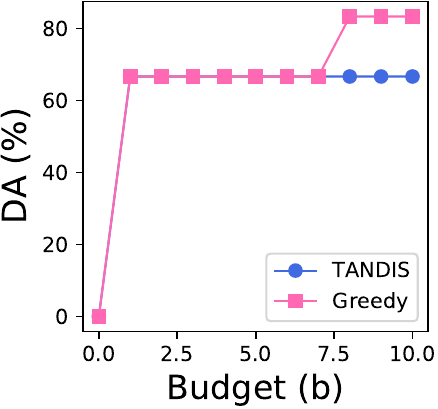}}\hfill
    \subfloat[Time (\cora)]{\includegraphics[width=0.23\textwidth]{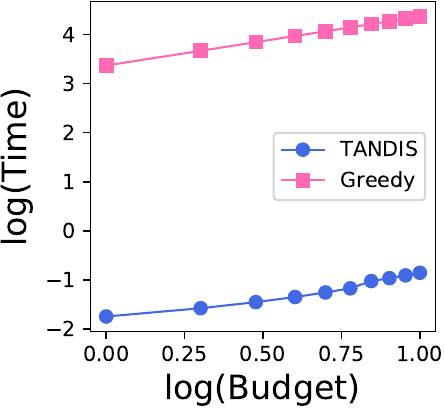}}
    \\
    \subfloat[\seer-\gcn]{\includegraphics[width=0.23\textwidth]{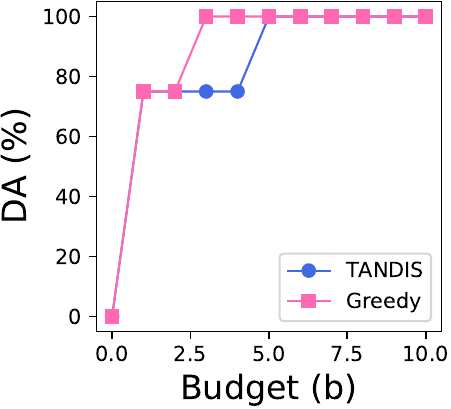}}\hfill
    \subfloat[\seer-\gsage]{\includegraphics[width=0.23\textwidth]{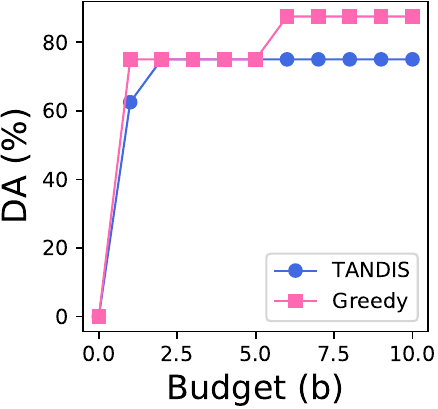}}\hfill
    \subfloat[\seer-\gat]{\includegraphics[width=0.23\textwidth]{Greedy/Citeseer/GAT.pdf}}\hfill
    \subfloat[Time (\seer)]{\includegraphics[width=0.23\textwidth]{Greedy/Citeseer/time.pdf}}
    \caption{\textbf{Comparison of \name with its non-neural Greedy equivalent for the node classification task in \cora and \seer averaged over $10$ targets. Greedy performs better than \name, but it takes time more than $4$ orders of magnitude with an improvement of only $\sim 20 \%$. So, Greedy is not practically feasible. }}
    \label{fig:greedy}
\end{figure*}

\subsubsection{Non-neural Baselines}
Here, we compare our method \name against two non-neural baselines --- \random and \degree, as defined in \cite{chang2020restricted}. Figure \ref{fig:non_neu_base} shows the results of these baselines for node classification task in \cora and \seer compared with \name. Note that our method comprehensively outperforms these baselines.



\subsubsection{Comparison with Greedy}

Figure ~\ref{fig:greedy} shows the comparison between our proposed RL optimization of Equation \ref{eq:new_obj} (i.e., \name) and the corresponding greedy optimization (i.e., Greedy) for $10$ targets on the node classification task in \cora and \seer. While Greedy does consistently better, the time taken is more than $4$ orders in magnitude (See Fig.~\ref{fig:greedy}(d) and Fig.~\ref{fig:greedy}(h). Thus, even though we can obtain better performance by greedily optimizing the proposed neighborhood distortion, the trade-off for efficiency is huge, which makes the proposed RL optimization more scalable and efficient for finding optimal perturbations maximizing neighborhood distortion (i.e., Equation \ref{eq:silh}).

\subsubsection{Interpretability}

In this section, we study additional node properties in order to interpret the neighborhood distortion. In order to do that, we choose the following different node-level properties.

\paragraph{Feature and Graph Properties}

\begin{itemize}
    \item \textbf{Feature Similarity}: Jaccard similarity of node features of the target and the candidate node for perturbation.
    
    \item \textbf{Degree}: In an undirected graph, the node degree is the number of edges adjacent to the node. In a directed graph, the in-degree and the out-degree of a node is the number of edges incident on the node, and the number of outgoing edges from the node respectively. In our case, since we have conducted our study on undirected graphs, the first definition holds true. However, a similar study can easily be conducted in case of a directed graph.
    
    \item \textbf{Local Clustering Coefficient}: The local clustering coefficient of a vertex is given by the ratio of the actual number of links between the vertices in its neighborhood divided and the total no. of possible links that can exist between these neighbor vertices. 
    
    \item \textbf{Reverse $k$-NN Rank}: Position of the candidate node in the ranked list of nodes, sorted in descending order, according to the number of times the node occurs in the 2-hop neighbourhood of all the other nodes in the graph. 
    
\end{itemize}

\paragraph{Community Based Properties}

We have also taken into consideration a few community related measures. For a given target $t$ we define its community, $S$, as the $50$ nearest neighbours in the original embedding space. Considering the fact that upon perturbation, the neighbourhood of the target becomes distorted, we define the perturbed community, $S'$, of target $t$ as the $50$ nearest neighbours in the perturbed embedding space. We expect that after perturbation, the quality of the community should worsen. Throughout the definitions stated further, m(S) denotes the number of edges between the nodes in community $S$, and c(S) denotes the number of edges between nodes in $S$ and the rest of the network, i.e., nodes in $\CV \backslash S$. c(S) is also known as cut size of $S$.

\begin{itemize}
    \item \textbf{Edge Expansion Difference ($\Delta EE$)}: The edge expansion ($EE$) is defined as the ratio of cut size and minimum of the cardinalities of the two communities $S$ and $\CV \backslash S$. We consider the edge expansion difference ($\Delta EE$) as:
    
    \begin{equation}
        \Delta EE = EE(S, \CV) - EE(S', \CV)
    \end{equation}
    where,
    \begin{equation}
        EE(S, \CV) = \frac{c(S)}{\min{\left(|S|,|\CV \backslash S|\right)}}
    \end{equation}
    
    denotes the edge expansion of  the community $S$.
    
    \item \textbf{Conductance Difference ($\Delta C$)}: Conductance ($C(S)$) is the fraction of edge volume that points outside the community $S$, where edge volume is defined as the sum of node degrees of the nodes in the community. We consider the conductance difference ($\Delta C$) as a property:
    
    \begin{equation}
        \Delta C = C(S) - C(S')
    \end{equation}
    where,
    \begin{equation}
        C(S) = \frac{c(S)}{\left(2m(S) + c(S)\right)}
    \end{equation}
    
    denotes the conductance of  the community $S$.

    \item \textbf{Volume Difference ($\Delta V$)}: The volume ($V(S)$) of a community $S$, is the sum of (out-)degrees of the nodes in $S$ in case the graph is directed, or only the sum of degrees of the nodes in $S$, if the graph is undirected. For our case the latter holds. We consider the volume difference ($\Delta V$) as a property:
    
    \begin{equation}
        \Delta V = V(S) - V(S')
    \end{equation}
    where,
    \begin{equation}
        V(S) = \left(2m(S) + c(S)\right),
    \end{equation}
    
    denotes the volume of  the community $S$.
    
    \item \textbf{Normalized Cut Size Difference}: The normalized cut size is the cut size multiplied with the reciprocal of the volumes of the two sets $S$ and $\CV \backslash S$.  We consider the normalized cut size difference ($\Delta NCS$) as a property:
    
    \begin{equation}
        \Delta NCS = NCS(S, \CV) - NCS(S', \CV)
    \end{equation}
    where,
    {\scriptsize
    \begin{equation}
        NCS(S, \CV) = c(S)\left(\frac{1}{2m(S) + c(S)} + \frac{1}{2(|\CE| + m(S)) - c(S)}\right),
    \end{equation}}
    
    denotes the normalized cut size of  the community $S$.
    
\end{itemize}

\begin{table}[t]
    \centering
    \resizebox{\linewidth}{!}{%
    \begin{tabular}{lrr}
    \toprule
    Property & Correlation & $p$-value \\
    \midrule
    \textbf{Feature similarity} & $\mathbf{-0.20 \pm 0.196}$ & $\mathbf{< 0.001}$ \\
    \textbf{Degree} & $\mathbf{-0.15 \pm 0.008}$ & $\mathbf{< 0.001}$ \\
    \textbf{Local Clustering Coefficient} & $\mathbf{0.25 \pm 0.004}$ & $\mathbf{< 0.001}$ \\
    Reverse k-NN rank & $0.02 \pm 0.008$ & 0.37 \\
    Edge Expansion Difference & $0.03 \pm 0.014$ & 0.24 \\
    Conductance Difference & $0.01 \pm 0.010$ & 0.65 \\
    Normalized-Cut Size Difference & $0.01 \pm 0.010$ & 0.67 \\
    Volume Difference & $0.02 \pm 0.017$  & 0.37 \\
    \bottomrule
    \end{tabular}}
    \caption{\textbf{Spearman's Rank Correlation of various node properties with neighborhood distortion along with their statistical significance.}}
\label{tab:interp_det}
\end{table}

Table \ref{tab:interp_det} shows the Spearman's Rank Correlation of these properties with the neighborhood distortion measure. As we have already seen in Section \ref{sec:detection}, only \textbf{Degree}, \textbf{Feature Similarity}, and \textbf{Local Clustering Coefficient} show significant correlations and the possible reasons for them. Now, it is important to note that the lack of significance in the correlation with the community-based metrics indicates that the neighborhood distortion may not necessarily bring a difference in the local community around the target node, as defined by these community based measures. In particular, the perturbation is not large enough to cause a significant change in the entire neighborhood community but only enough to cause a change in the inference for the downstream task. Additionally, the low magnitude of significant correlation with the three parameters (i.e., degree, feature similarity, and local clustering coefficient) further strengthens the claim that the distortion is difficult to detect with standard attack detection techniques \cite{survey}. 


\subsection{Reproducibility and Code-base}

The code is made available at \url{https://github.com/idea-iitd/TANDIS}.